\documentclass[lettersize,journal]{IEEEtran}
\usepackage{amsmath,amsfonts,bm}
\usepackage{algorithmic}
\usepackage{algorithm}
\usepackage{array}
\usepackage{multirow,subfigure,caption,amssymb,booktabs,xcolor}
\usepackage[caption=false,font=normalsize,labelfont=sf,textfont=sf]{subfig}
\usepackage{textcomp}
\usepackage{stfloats}
\usepackage{url}
\usepackage{verbatim}
\usepackage{graphicx}
\usepackage{cite}
\newcommand{\modelname}{VLS-GPT}

\usepackage{amsthm}
\usepackage{xcolor}
\newtheorem{prop}{Proposition}

\newcommand{\revfirst}{\textcolor{black}}
\newcommand{\revsecond}{\textcolor{black}}

\begin{document}

\title {Variational Latent-State GPT for Semi-Supervised Task-Oriented Dialog Systems}  
\author {
    Hong Liu,
    Yucheng Cai,
    Zhenru Lin,
    Zhijian Ou,~\IEEEmembership{Senior Member, IEEE},
    Yi Huang,
    Junlan Feng,~\IEEEmembership{Fellow, IEEE}
    \thanks{Hong Liu, Yucheng Cai, Zhenru Lin, Zhijian Ou are with Speech Processing and Machine Intelligence Lab, Tsinghua University, Beijing, China (e-mails: \{liuhong21, caiyc18, linzr18\}@mails.tsinghua.edu.cn; ozj@tsinghua.edu.cn).
    Yi Huang, Junlan Feng are with China Mobile Research Institute, Beijing, China (e-mails: \{huangyi, fengjunlan\}@chinamobile.com).
    All the authors are also affiliated with Tsinghua University-China Mobile Communications Group Co., Ltd. Joint Institute, Beijing, China.
     (Corresponding author: Zhijian Ou)}
}

\markboth{IEEE/ACM TRANSACTIONS ON AUDIO, SPEECH AND LANGUAGE PROCESSING, 2022}%
{Shell \MakeLowercase{\textit{et al.}}: A Sample Article Using IEEEtran.cls for IEEE Journals}


\maketitle

\begin{abstract}
Recently, two approaches, fine-tuning large pre-trained language models and variational training, have attracted significant interests, separately, for semi-supervised end-to-end task-oriented dialog (TOD) systems.
In this paper, we propose Variational Latent-State GPT model (VLS-GPT), which is the first to combine the strengths of the two approaches.
Among many options of models, we propose the generative model and the inference model for variational learning of the end-to-end TOD system, both as auto-regressive language models based on GPT-2, which can be further trained over a mix of labeled and unlabeled dialog data in a semi-supervised manner.
Variational training of VLS-GPT is both statistically and computationally more challenging than previous variational learning works for sequential latent variable models, which use turn-level first-order Markovian. The inference model in VLS-GPT is non-Markovian due to the use of the Transformer architecture. 
In this work, we establish Recursive Monte Carlo Approximation (RMCA) to the variational objective with non-Markovian inference model and prove its unbiasedness. Further, we develop the computational strategy of sampling-then-forward-computation to realize RMCA, which successfully overcomes the memory explosion issue of using GPT in variational learning and speeds up training.
Semi-supervised TOD experiments are conducted on two benchmark multi-domain datasets of different languages - MultiWOZ2.1 and CrossWOZ. VLS-GPT is shown to significantly outperform both supervised-only and semi-supervised self-training baselines. 
\end{abstract}


\section{Introduction}
\begin{figure}[t]
\centering
	\includegraphics[width=0.95\linewidth]{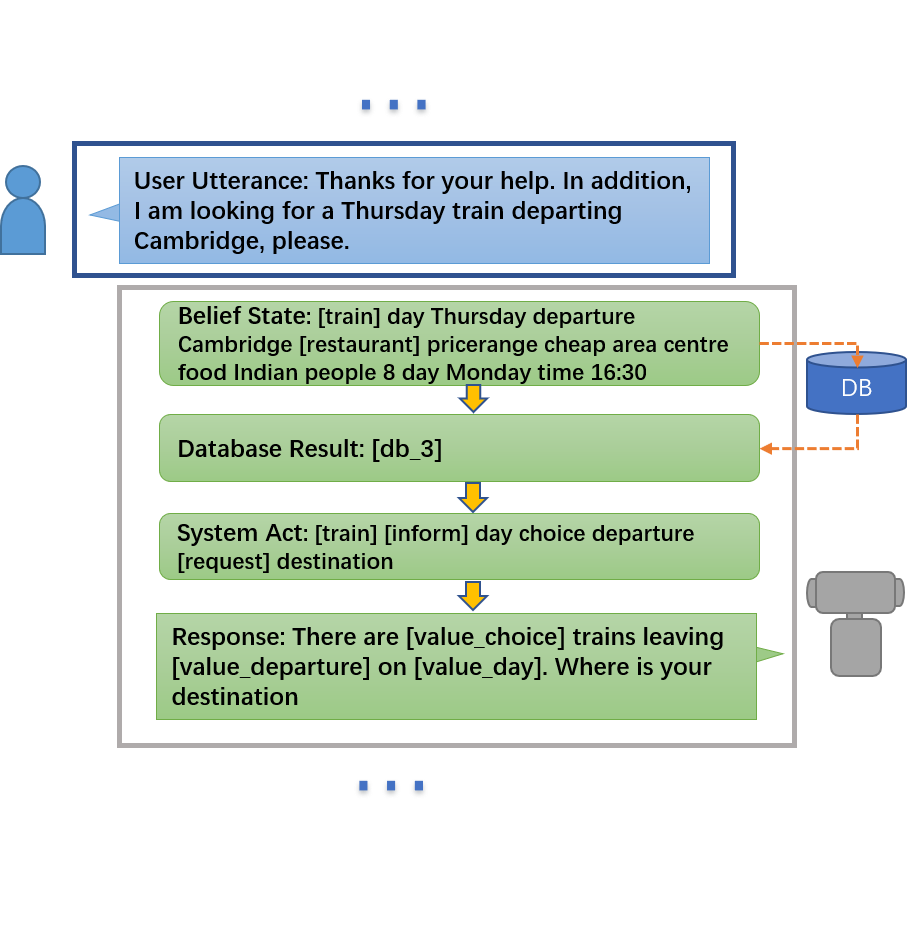}
	\vspace{-1.5em}
	\caption{The information flow in one turn from a task-oriented dialog. Square brackets denote special tokens in GPT-2.}
	\vspace{-0.5em}
	\label{fig:flow}
\end{figure}

Task-oriented dialogue (TOD) systems are mainly designed to assist users to accomplish their goals, which usually consists of several modules for tracking user goals (often called the belief states), querying a task-related database (DB), deciding actions and generating responses.
The information flow in a task-oriented dialog is illustrated in Figure~\ref{fig:flow}, which involves user utterances, belief states, DB results, system acts and responses.
The methodology for building TOD systems is gradually advancing from separate training of individual modules \cite{mrkvsic2017neural, wen2017latent} to the end-to-end (E2E) trainable approach \cite{wen2017a, liu2017end, lei2018sequicity, fsdm, zhang2020task, gao2020paraphrase}.
E2E methods usually employ the encoder-decoder seq2seq architecture \cite{sutskever2014sequence} to connect modules and train them together.
Incorporating intermediate supervisions from \emph{annotated} belief states and system acts, and optimizing the system jointly for belief state tracking, action and response generation in multi-task settings, is found to significantly improve the performance \cite{lei2018sequicity, fsdm, zhang2020task}. 

Although E2E methods have achieved promising results, they usually require substantial amounts of domain-specific manually labeled data. The long-standing labeled-data scarcity challenge, which hinders efficient development of TOD systems at scale, is even magnified in building E2E TOD systems. There are increasing interests in developing semi-supervised learning (SSL) \cite{zhu2006semi} methods for E2E TOD systems, which aims to leverage
both labeled and unlabeled data. Remarkably, two SSL approaches have attracted significant interests for semi-supervised E2E TOD systems.

First, a broad class of SSL methods formulates a latent variable model (LVM) of observations and labels and blends unsupervised and supervised learning \cite{zhu2006semi}. Unsupervised learning with LVM usually maximizes the marginal likelihood via variational learning \cite{kingma2013auto}.
This approach has been studied \cite{sedst, zhang-etal-2020-probabilistic} for semi-supervised TOD systems, and the models typically use LSTM based seq2seq architectures.
Another broad class of SSL methods is unsupervised pre-training, where the goal is to find a good initialization point instead of modifying the supervised learning objective \cite{radford2018improving}.
In the pre-training-and-fine-tuning approach, large-scale language models pre-trained on open-domain texts, such as BERT (Bidirectional Encoder Representations from Transformers) \cite{devlin2019bert}, GPT (Generative Pre-Training) \cite{radford2018improving}), are fine-tuned with in-domain labels \cite{heck2020trippy,budzianowski-vulic-2019-hello}. 
Particularly, Transformer \cite{vaswani2017attention} based auto-regressive language models, like GPT-2 \cite{radford2019gpt2}, learn a strong distribution for next-token prediction, which makes them particularly useful for generative TOD systems \cite{budzianowski-vulic-2019-hello, ham-etal-2020-end, hosseini2020simple, peng2020etal, kulhanek2021augpt, yang2021ubar}.

Remarkably, the two approaches, pre-training-and-fine-tuning and LVM based variational training, are not mutually exclusive and could be jointly used, and conceivably, can complement each other. The pre-training approach is powerful at leveraging unlabeled open-domain data, while the variational approach is suited to exploiting unlabeled in-domain data\footnote{Variational semi-supervised learning with LVM generally assumes that the unlabeled and labeled data are drawn from the same distribution, except that the unlabeled data are missing data (without labels) \cite{kingma2013auto}. This is often occurred in real-world situations, e.g. unlabeled in-domain data are easily available between customers and human agents.}. 
Particularly, both applications of pre-trained GPT and variational learning are previously known separately in the literature for semi-supervised TOD systems.
\emph{But how we can leverage both pre-trained GPT and variational learning is not clear, presents new challenges and has not ever been examined.}

\begin{figure}[t!]
	\centering
	\subfigure[Generative Model]
	{
		\includegraphics[width=0.95\columnwidth]{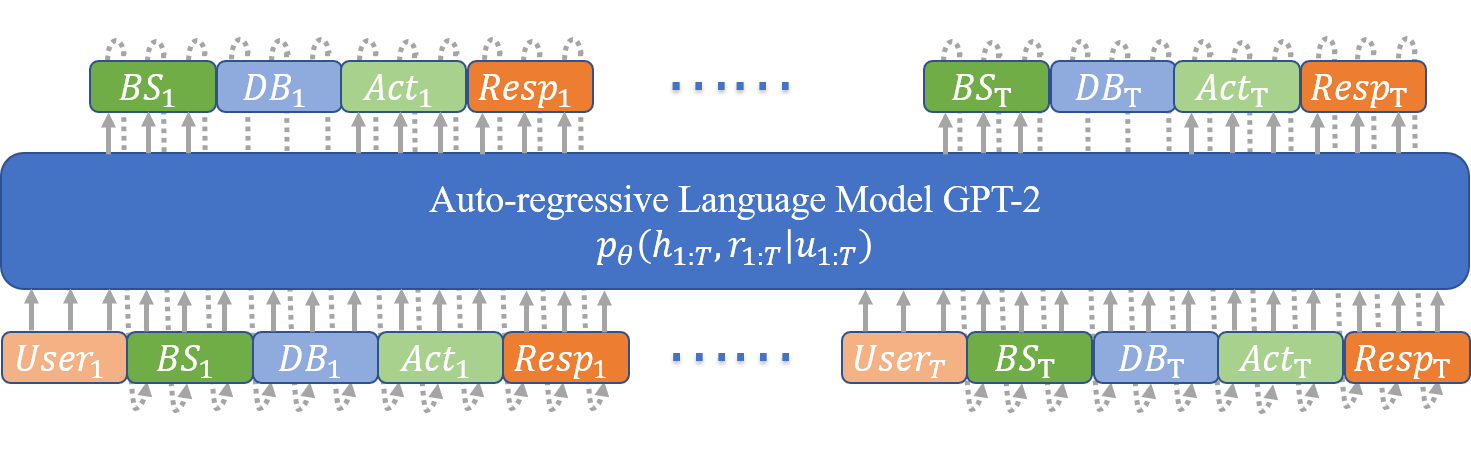}}
	    \label{generative}
	    \subfigure[Inference Model]
	{
		\includegraphics[width=0.95\columnwidth]{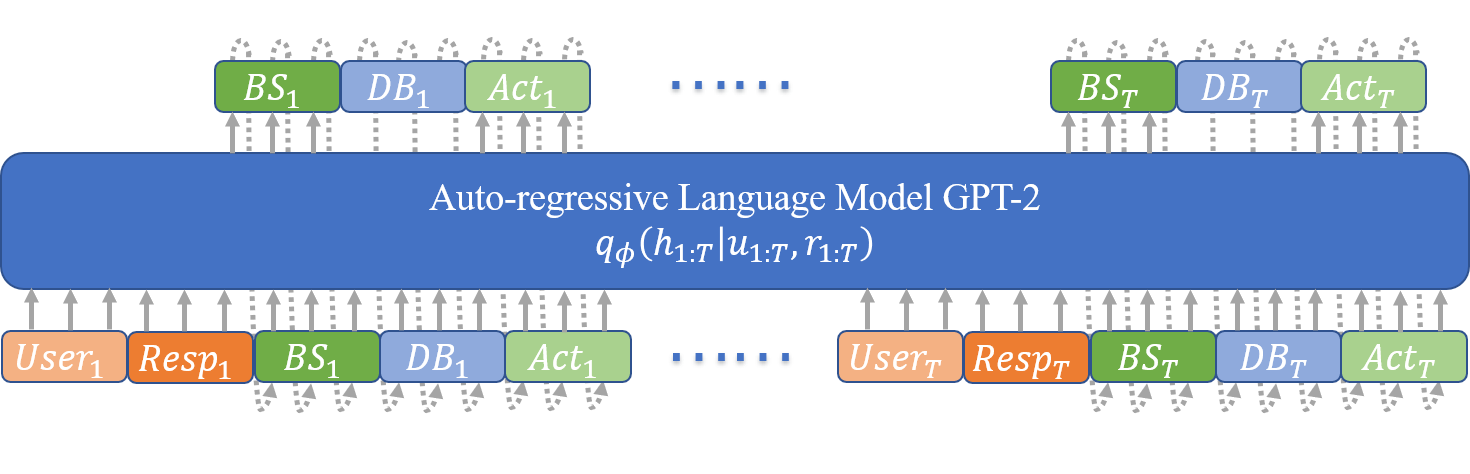}}
	    \caption{An overview of \modelname{}, which consists of two auto-regressive lanugage models - a generative model and an inference model, both initialized from GPT-2 but trained with different training sequences as shown in Figure \ref{fig:examples}.}
	    \label{inference}
	\label{structure}
\end{figure}

To answer the aforementioned question, we develop Variational Latent-State GPT model (\modelname{}), which successfully combines the pretraining and variational approaches for semi-supervised TOD Systems.
Among many options of models, we propose the \emph{generative model} and the \emph{inference model} for variational learning of the end-to-end TOD system, both as auto-regressive language models based on GPT-2, as shown in Fig. \ref{structure}.
To be clear, GPT-2 \cite{radford2019gpt2} in this paper refers to the particular class of causal language models, which computes conditional probabilities for next-token generation via self-attention based Transformer neural network \cite{vaswani2017attention}.

\modelname{} takes all the intermediate states (including the belief states, DB results and system acts) as latent variables. The generative model iteratively generates belief states, DB results, system acts and response given user inputs, and the inference model iteratively infers all intermediate states given user inputs and system responses.
Both the generative model and the inference model are initialized from the pretrained GPT-2, and can be further trained (finetuned) over a mix of labeled and unlabeled in-domain dialog data from the targeted task in a semi-supervised manner.
Semi-supervised TOD experiments are conducted on two benchmark multi-domain datasets of different languages, MultiWOZ2.1 \cite{eric2019multiwoz} and CrossWOZ \cite{zhu2020crosswoz}, which are in English and Chinese respectively. \modelname{} is shown to significantly outperform both supervised-only and semi-supervised self-training baselines.

\modelname{} builds on prior work on using pretrained GPT and variational learning for semi-supervised TOD systems, and makes the following contributions in model, algorithm, and experiment, respectively.
\begin{itemize}
\item \modelname{} is the first to combine the strengths of large pre-trained language model and variational learning for semi-supervised TOD systems.
Previous GPT based TOD systems, e.g. SimpleTOD \cite{hosseini2020simple} and UBAR \cite{yang2021ubar}, only conduct supervised learning. LABES \cite{zhang-etal-2020-probabilistic} employs variational learning, but only uses turn level LSTM based generative and inference models.
\item Variational training of VLS-GPT is both statistically and computationally more challenging than previous variational learning works for sequential latent variable models \cite{kim2020sequential,zhang-etal-2020-probabilistic}, which use turn-level first-order Markovian. The inference model in VLS-GPT is non-Markovian due to the use of the Transformer architecture. 
In this work, we establish Recursive Monte Carlo Approximation (RMCA) to the variational objective with non-Markovian inference model and prove its unbiasedness. Further, we develop the computational strategy of sampling-then-forward-computation to realize RMCA, which successfully overcomes the memory explosion issue of using GPT in variational learning and speeds up training.
\item We conduct extensive experiments on two benchmark multi-domain datasets of different languages (MultiWOZ2.1 in English and CrossWOZ in Chinese) and demonstrate the effectiveness of \modelname{} in semi-supervised TOD experiments, outperforming both supervised-only and semi-supervised baselines.
Overall, VLS-GPT using 50\% labels can obtain close performance to the strong GPT-based supervised-only baseline on 100\% labeled data. \revsecond{We release the code to reproduce our experiments at \url{https://github.com/thu-spmi/VLS-GPT}}.
\end{itemize}

\begin{table*}[t]
    \caption{Comparison of existing GPT-based TOD methods by their training objectives. 
		LABES is also shown to compare with VLS-GPT. For LABES and \modelname{}, we show objectives for training their generative models.
		$u_t, b_t, d_t, a_t, r_t$, denote user utterance, belief state, DB result, system act, and response, respectively, for dialog turn $t$ in a dialog of $T$ turns.
		The subscript operates on each element in the bracket, e.g. $\{b,d,r,t\}_t$ is a shorthand for $b_t,d_t,a_t,r_t$.
	}
	\centering
	\resizebox{0.99\linewidth}{!}{
		\begin{tabular}{cl}
			\toprule
			
			Model &Training Objective\\
			\midrule
			B\&V \cite{budzianowski-vulic-2019-hello} &$ \prod_{t=1}^{T}p(r_t | \{u, b, d\}_t)$\\
			Ham et al. \cite{ham-etal-2020-end} & $\prod_{t=1}^{T}p( \{b, a, r\}_t|\{u, r\}_1, \cdots, \{u,r\}_{t-1}, u_t)$\\
			SOLOIST, AuGPT & $\prod_{t=1}^{T}p( \{b, d, r\}_t|\{u, r\}_1, \cdots, \{u,r\}_{t-1}, u_t)$\\
			SimpleTOD & $\prod_{t=1}^{T}p(\{u, r\}_1, \cdots, \{u,r\}_{t-1}, \{u, b, d, a, r\}_t)$\\
			LABES & $\prod_{t=1}^{T} p(\{b, d, r\}_t|r_{t-1}, b_{t-1}, u_t)$\\ 
			UBAR & $p(\{u, b, d, a, r\}_1, \cdots, \{u, b, d, a, r\}_T) = \prod_{t=1}^{T} p(\{u, b, d, a, r\}_t|\{u, b, d, a, r\}_1, \cdots, \{u, b, d, a, r\}_{t-1}) $\\
			\modelname{} & $p(\{b, d, a, r\}_1, \cdots, \{b, d, a, r\}_T|u_1, \cdots, u_T) = \prod_{t=1}^{T} p(\{b, d, a, r\}_t|\{u, b, d, a, r\}_1, \cdots, \{u, b, d, a, r\}_{t-1}, u_t)$\\
			\bottomrule
	    \end{tabular}}
	\label{comparison}
\end{table*}

\section{Related Work}
\label{related}
\subsection{Semi-supervised TOD systems with pre-trained GPT-2}

GPT-2 is an auto-regressive language model (LM), pre-trained over large amounts of open-domain data, which can be fine-tuned to accomplish a range of natural language processing tasks.
The pre-training-and-fine-tuning approach broadly falls under the category of semi-supervised learning \cite{radford2018improving}. 
Two early studies in finetuning GPT-2 on labeled dialog data for TOD systems are \cite{budzianowski-vulic-2019-hello} and \cite{ham-etal-2020-end}.
Later, two similar further developments are proposed, namely SimpleTOD \cite{hosseini2020simple} and SOLOIST \cite{peng2020etal}.
Two recent studies are AuGPT \cite{kulhanek2021augpt} and UBAR \cite{yang2021ubar}.
AuGPT proposes a modification of the loss function and a data augmentation strategy based on back-translation. 
UBAR proposes the session-level finetuning of GPT-2, namely on the whole sequence of the entire dialog session which is composed of user utterances, belief states, DB results, system acts and responses of all dialog turns. This is different from the turn-level training, employed in all previous works. Moreover, UBAR also performs session-level evaluation, which means it uses previous generated responses instead of the ground truth to form the context for current turn. We summarize the differences between existing GPT-based TOD methods by their training objectives in Table \ref{comparison}. Notably, all previous GPT-based TOD systems only conduct supervised learning of the generative model.

\modelname{} adopts the session-level training and evaluation as in UBAR, which is found to be useful.
But as can be seen from Table \ref{comparison}, \modelname{} uses a new objective for training the generative model, which is different from that used in UBAR. \modelname{} does not calculate the cross-entropy loss over user utterances, while UBAR does.
This is important for VLS-GPT in developing variational learning, since both the generative model and the inference model in VLS-GPT are defined as conditional distributions given user utterances for variational learning.

\subsection{Semi-supervised TOD systems with variational latent variable models.}
\label{sec:variational-latent}
Variational latent variable models have been used in TOD systems \revfirst{with two different, orthogonal aims.
In the first class of studies, latent variables are introduced to model the system acts of a TOD system, which aims to help reinforcement learning (RL) of dialog policy.
The second aim, which is also the aim of this work, is to enable semi-supervised training of a TOD system, where belief states (optionally with other annotations) are treated as latent variables.
Notably, two fundamental abilities of a TOD system are tracking of the belief states and planning of the system actions \cite{lubis2020lava}.
It is interesting to see that the two classes of studies aim to enhance the two fundamental abilities of a TOD system respectively.
}

\revfirst{
For the first class of modeling system acts, typical studies include LIDM \cite{wen2017latent}, LaRL \cite{zhao2019rethinking}, and LAVA \cite{lubis2020lava}.
Traditional approaches use handcrafted system acts.
LIDM \cite{wen2017latent} employs a categorical latent variables to discover dialog intentions (i.e. system acts), which is similar to unsupervised clustering.
LaRL \cite{zhao2019rethinking} and LAVA \cite{lubis2020lava} follows the latent action framework and uses the latent space of a variational model as the action space.
The motivation is to alleviate the problem of large action spaces and long trajectories of word-level RL (i.e. using the entire output vocabulary as the action space), instead of towards semi-supervised learning of TOD systems.
On top of LaRL, LAVA \cite{lubis2020lava} further leverages auxiliary tasks to shape the latent variable distribution to yield a more action-characterized latent representation.
}
Recently, PLATO \cite{bao-etal-2020-plato} also uses a $K$-way categorical latent variable, still modeling system actions, to tackle the inherent one-to-many mapping problem in response generation.

For the second class, there are previous studies in using latent variable models for semi-supervised TOD systems. SEDST \cite{sedst} uses a combination of posterior regularization and auto-encoding to perform semi-supervised learning for belief tracking. 
LABES \cite{zhang-etal-2020-probabilistic} is an inspiring related work, which models belief states as latent variables and employs variational learning. However, only turn-level LSTM based generative and inference models are used in LABES; In contrast, \modelname{} adopts session-level GPT based models. Such difference can be seen from Table \ref{comparison} for the generative models. 
Correspondingly, the session-level inference model designed in this paper for VLS-GPT is radically different from that in LABES, which is non-Markovian, and we need to address new challenges in using GPT in variational learning, both statistically and computationally.
To the best of our knowledge, combining both pre-trained GPT and variational learning for semi-supervised TOD systems has not been explored yet.

\section{preliminaries}
\subsection{Variational learning} \label{sec:vae}
Here we briefly review the variational learning methods, recently developed for learning latent variable models \cite{kingma2013auto,Rezende2014StochasticBA}.
Consider a latent variable model $p_\theta(x,z)$ for observation $x$ and latent variable $z$, with parameter $\theta$.
Instead of directly maximizing the marginal log-likelihood $\log p_\theta(x)$ for the above latent variable model, variational methods maximize the following variational evidence lower bound (ELBO), after introducing an auxiliary inference model $q_\phi(z|x)$ to approximate the true posterior $p_\theta(z|x)$:
\begin{displaymath}
ELBO(\theta,\phi;x) \triangleq \mathbb{E}_{q_\phi(z|x)} \left[ \log \frac{p_\theta(x,z)}{q_\phi(z|x)} \right]
\end{displaymath}
It is known that the gradient of ELBO with respect to (w.r.t.) $\theta$ can be reliably estimated with a single Monte Carlo sample:
\begin{displaymath}
\frac{\partial}{\partial \theta}ELBO(\theta,\phi;x) \approx \frac{\partial}{\partial \theta} \log p_\theta(x,{z}),  {z} \sim q_\phi(z|x)
\end{displaymath}
Estimating the gradient of ELBO w.r.t. $\phi$ in the case of continuous $z$ can be effectively performed via the reparameterization trick \cite{kingma2013auto,Rezende2014StochasticBA}, but is challenging for the case of discrete $z$, mainly due to the difficulty in estimating the second term:
\begin{equation} \label{eq:ELBO}
\begin{split}
&\frac{\partial}{\partial \phi}ELBO(\theta,\phi;x)\\
= & \mathbb{E}_{q_\phi(z|x)} \left[ \frac{\partial}{\partial \phi} \log \frac{p_\theta(x,z)}{q_\phi(z|x)} \right]
+\sum_z \left[\frac{\partial}{\partial \phi} q_\phi(z|x)\right] \log \frac{p_\theta(x,z)}{q_\phi(z|x)}
\end{split}
\end{equation}

For estimating gradients with discrete latent variables, some methods have been proposed, as reviewed in \cite{ou2018review}.
The classic REINFORCE trick \cite{williams1992simple} can suffer from high variance, and various variance reduction techniques have been developed to make the estimator more usable. The categorical reparameterization trick \cite{jang2016categorical} relaxes discrete variables to be continuous variables computed by the Gumbel-Softmax function and then apply the reparameterization trick to estimate the gradients.

\subsection{The Straight-Through trick} \label{sec:STT}
For scenarios in which we need to sample discrete values (e.g. from a vocabulary of tokens) in addition to estimating the gradients, the Straight-Through \cite{bengio2013estimating} gradient estimator is attractive.
To study the estimation of the second term in Eq.(\ref{eq:ELBO}), we consider the illustrative problem of estimating the gradient of the expectation of $f(z)$ where $z$ is a discrete variable with distribution $q_\phi(z)$ over the domain $\{1,2,\cdots,K\}$, i.e.
\begin{equation} \label{eq:discrete-grad}
\frac{\partial}{\partial \phi} \mathbb{E}_{q_\phi(z)} \left[ f(z) \right] = \frac{\partial}{\partial \phi} \sum_{z=1}^K q_\phi(z) f(z)
\end{equation}
Denote $\mathbf{z} = onehot(z)$ by encoding $z$ as the $K$-dimensional one-hot vector and hereafter we can rewrite $f(z)$ as $f(\mathbf{z})$ by abuse of notation.
Assume the probability vector $\bm{\pi} = (q_\phi(1),q_\phi(2),\cdots,q_\phi(K))$ is denoted shortly as $\bm{\pi}$, which is usually calculated by softmax function on top of neural networks parameterized by $\phi$.
Here we suppress the dependence of $\bm{\pi}$ on $\phi$ to reduce notational clutter.

The basic idea of the Straight-Through gradient estimator is that the sampled discrete values are used for forward computation, and the continuous softmax probabilities are used for backward gradient calculation\footnote{The Straight-Through trick can be used in combination with Gumbel-Softmax \cite{jang2016categorical}, called Straight-Through Gumbel-Softmax estimator, which can tune a temperature hyper-parameter to balance estimator bias and variance.
We find the Straight-Through estimator works pretty well in our experiments, and leave the exploration of other estimators as future work.}. 
Specifically, the gradient in Eq.(\ref{eq:discrete-grad}) is approximated with a single Monte Carlo sample ${z} \sim q_\phi(z)$, as follows:
\begin{equation} \label{eq:st-grad}
\frac{\partial}{\partial \phi} \mathbb{E}_{q_\phi(z)} \left[ f(z) \right] \approx \frac{\partial f(\mathbf{z})}{\partial \mathbf{z}} \frac{\partial {\mathbf{z}}}{\partial \phi} \approx \frac{\partial f(\mathbf{z})}{\partial \mathbf{z}} \frac{\partial {\bm{\pi}}}{\partial \phi}
\end{equation}

It can be seen that the above Straight-Through Trick (STT) can be realized by representing the one-hot vector of each discrete variable $z$ as follows, whenever feeding $z$ forward:
\begin{equation}\label{eq:ST-trick}
STT(z)=\mathbf{z}+\bm{\pi}-\bm{\pi}.detach
\end{equation}
where $\bm{\pi}.detach$ means that we do not calculate its gradient during back-propagation.
It can be seen that applying the above $STT(z)$ in the forward direction and computing back-propagation as usual realizes the Straight-Through gradient estimator Eq.(\ref{eq:st-grad}), and thus successfully propagates gradients through $z$ in the backward direction.

\section{Method}
In the following, we first introduce the \modelname{} model, as shown in Fig. \ref{structure}, then we describe the supervised learning and semi-supervised learning methods based on \modelname{}, respectively. Finally, we elaborate on the statistical and computational strategies, which enables us to perform variational training for the GPT-2 based models.

\subsection{Model}
\textbf{Notations~}
Consider the information flow in a task-oriented dialog of $T$ turns, as illustrated in Figure~\ref{fig:flow}, and let $u_t$ denote the user utterance, $b_t$ the belief state, $d_t$ the database result, $a_t$ the system action and $r_t$ be the delexicalized response, respectively, at turn $t=1,\cdots,T$, which all are represented as token sequences.
Denote the token sequence, for example, for $h_t$ by $h_{t}^{(i)}, i=1,\cdots,|h_t|$, where $|h_t|$ denotes the length of $h_t$ in tokens.
The vocabulary size of tokens is $K$.
Denote the sub-sequence $h_1,\cdots,h_{t-1}$ by $h_{<t}$, similarly $h_{t}^{(<i)}$ for $h_{t}^{(1)},\cdots,h_{t}^{(i-1)}$.

Motivated by recent studies \cite{hosseini2020simple,yang2021ubar}, we unify the workflow of a TOD system (belief state tracking, action and response generation) into a single sequence prediction problem, which can be accomplished by an auto-regressive language model.
In this work, the auto-regressive model for dialog generation is denoted by the conditional distribution
$p(\{b, d, a, r\}_1, \cdots, \{b, d, a, r\}_T|u_1, \cdots, u_T)$ as described in Table \ref{comparison}.
Given user utterances $u_{1:T}$, the belief states, DB results, system actions and responses $b_{1:T},d_{1:T},a_{1:T},r_{1:T}$ are recursively generated\footnote{The DB results $d_{1:T}$ are obtained by querying the database using the generated belief states.} according to $p_\theta(b_{1:T},d_{1:T},a_{1:T}, r_{1:T}|u_{1:T})$.
Specifically, at the first turn $t=1$, given $u_1$, the model sequentially generates $b_1,d_1,a_1,r_1$. At turn $t$, based on all previous user utterances and all generated outputs $u_1, b_1,d_1,a_1,r_1, \cdots, u_{t-1},b_{t-1},d_{t-1},a_{t-1},r_{t-1}$ and current user utterance $u_t$, the model sequentially generates $b_t,d_t,a_t,r_t$.
It can be easily seen that such recursive generation completes the entire dialog session.

A shorthand for $p(\{b, d, a, r\}_1, \cdots, \{b, d, a, r\}_T|u_1, \cdots, u_T)$ is $p_\theta(b_{1:T},d_{1:T},a_{1:T}, r_{1:T}|u_{1:T})$, and further will be written as $p_\theta(h_{1:T}, r_{1:T}|u_{1:T})$ for brevity. $h_t=\{b_t, d_t, a_t \}$ denotes the concatenation of intermediate states, which are observed in labeled dialogs, but become latent variables in unlabeled dialogs.
Note that these are simplified notations, which should obey the auto-regressive dialog generation, as explained above.
Further, the generative model can be decomposed as:
\begin{align} \label{eq:p_theta}
&p_\theta(h_{1:T}, r_{1:T}|u_{1:T})\\
=&\Pi_{t=1}^{T} p_\theta(h_t|\{u, h, r\}_1, \cdots, \{u, h, r\}_{t-1}, u_{t}) \nonumber \\
&~~~~~~~\times p_\theta(r_t|\{u, h, r\}_1, \cdots, \{u, h, r\}_{t-1}, \{u,h\}_{t}) \nonumber\\
\triangleq &\Pi_{t=1}^{T} p_\theta(h_t| h_{<t},r_{<t})  p_\theta(r_t|h_{<t},r_{<t},h_t) \nonumber
\end{align}
where, intuitively, we refer the conditional distribution $p_\theta(h_{t}|h_{<t},r_{<t})$ as the latent state prior, and $p_\theta(r_{t}|h_{<t},r_{<t},h_t)$ the response probability.
To reduce notational clutter, we suppress the conditioning of $h_t$ on user utterances in $p_\theta(h_t| h_{<t},r_{<t})$, which actually should follow the auto-regressive generation structure as emphasized above.
Similarly for the notation $p_\theta(r_t|h_{<t},r_{<t},h_t)$.

In order to perform unsupervised variational learning from unlabled dialogs (to be detailed below), we need an inference model $q_\phi(h_{1:T}|u_{1:T},r_{1:T})$ to approximate the true posterior $p_\theta(h_{1:T}|u_{1:T},r_{1:T})$, which is defined as follows:
\begin{align}\label{eq:q_phi}
&q_\phi(h_{1:T}|u_{1:T},r_{1:T})\\
=&\Pi_{t=1}^{T} q_\phi(h_t|\{u, r, h\}_1, \cdots, \{u, r, h\}_{t-1}, \{u,r\}_{t}) \nonumber\\
\triangleq&\Pi_{t=1}^{T} q_\phi(h_t|h_{<t},r_{<t},r_t) \nonumber
\end{align}
where similarly we suppress the conditioning of $h_t$ on user utterances in the auto-regressive inference structure as shown in Eq.(\ref{posterior}) below.

The \modelname{} model thus consists of two auto-regressive models - the generative model $p_\theta(h_{1:T}, r_{1:T}|u_{1:T})$ and the inference model $q_\phi(h_{1:T}|u_{1:T},r_{1:T})$, both initialized from GPT-2 but structured to be trained with different training sequences, as described below. 
\revfirst{The two models in \modelname{} are denoted by \modelname{}-p and \modelname{}-q respectively}.

\subsection{Supervised learning} 
\begin{figure}[t]
	\centering
	\subfigure[Training sequence for the generative model]
	{\includegraphics[width=0.99\columnwidth]{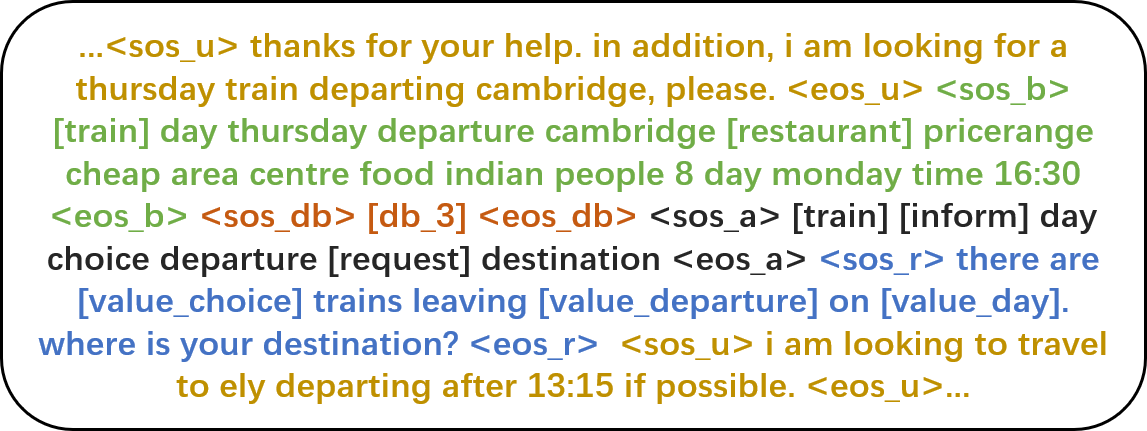}}
	\quad
	\subfigure[Training sequence for the inference model]
	{\includegraphics[width=0.99\columnwidth]{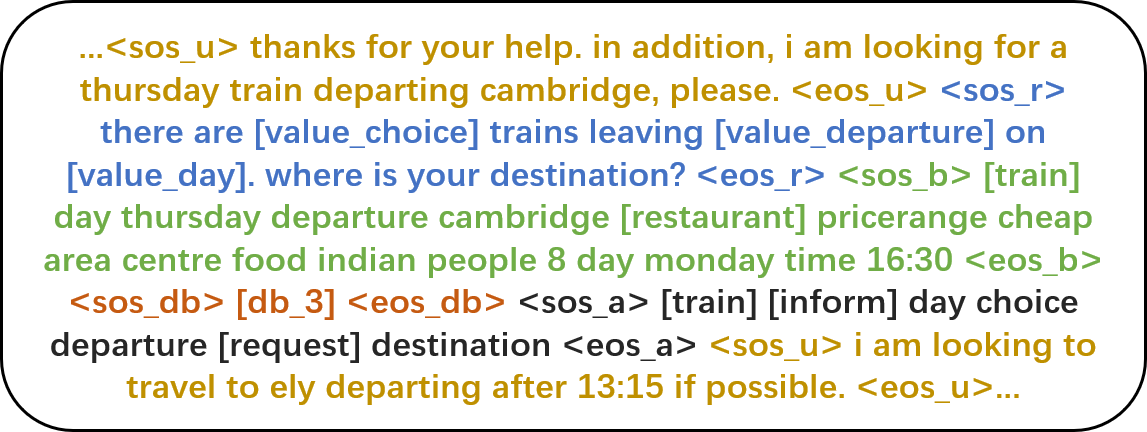}}
	\caption{Examples of training sequences described in Eq. (\ref{prior}) and Eq. (\ref{posterior}). Note that a complete training sequence contains many turns concatenated together.}
	\label{fig:examples}
\end{figure}

In supervised learning, the entire dialog is labeled. The training sequence for the generative model \revfirst{\modelname{}-p} is obtained by the concatenation as follows\footnote{The training sequence for the generative model VLS-GPT-p is the same as in UBAR. But as shown in Table \ref{comparison}, the training objective in VLS-GPT-p is $p_\theta(h_{1:T}, r_{1:T}|u_{1:T})$, which is different from UBAR and brings minor performance improvement as shown in Table \ref{sup-results}.}:
\begin{equation}\label{prior}
{u_1, b_1, d_1, a_1, r_1,..., u_T, b_T, d_T, a_T, r_T}
\end{equation}
And the training sequence for the inference model \revfirst{\modelname{}-q} is organized as:    
\begin{equation}\label{posterior}
{u_1, r_1, b_1, d_1, a_1,..., u_T, r_T, b_T, d_T, a_T}
\end{equation}
See examples in Figure \ref{fig:examples}.
Both models can then be trained from these training sequences through maximizing their likelihoods $p_\theta(h_{1:T}, r_{1:T}|u_{1:T})$ and $q_\phi(h_{1:T}|u_{1:T},r_{1:T})$ respectively, via teacher-forcing.

\subsection{Semi-supervised learning}
\label{sec:VL_learning}
When a mix of labeled and unlabeled data is available, we perform semi-supervised learning, which essentially is a combination of supervised learning and unsupervised variational learning \cite{zhu2006semi,kingma2013auto}.
Specifically, we first conduct supervised pre-training of \modelname{} on labeled data. Then we alternately draw supervised and unsupervised mini-batches from labeled and unlabeled data, and update the generative model and the inference model via supervised gradients and unsupervised gradients, respectively.
The supervise gradients are calculated the same as in supervised learning.

For unsupervised learning, the intermediate states $b_{1:T}, d_{1:T}$ and $a_{1:T}$ (simply $h_{1:T}$) are unlabeled. Thus, we maximize marginal likelihood, which is translated to maximizing the following variational bound (ELBO):
\begin{displaymath} 
\mathcal{J}_{\text{VL}}
=\mathbb{E}_{q_\phi(h_{1:T}|u_{1:T},r_{1:T})}\left[\log \frac{p_\theta(h_{1:T},r_{1:T}|u_{1:T})}{q_\phi(h_{1:T}|u_{1:T},r_{1:T})} \right]
\end{displaymath}

Plugging the GPT-based generative and inference models (Eq. (\ref{eq:p_theta}) and (\ref{eq:q_phi})) into the above ELBO objective function, we obtain
\begin{equation} \label{eq:J_VL_2}
\begin{split}
\mathcal{J}_{\text{VL}}
=&\mathbb{E}_{q_\phi(h_{1:T}|u_{1:T},r_{1:T})}\left[\sum_{t=1}^{T} \log p_\theta(r_{t}|h_{<t},r_{<t},h_t)
\right]\\
&+\mathbb{E}_{q_\phi(h_{1:T}|u_{1:T},r_{1:T})}\left[\sum_{t=1}^{T} \log \frac{p_\theta(h_{t}|h_{<t},r_{<t})}{q_\phi(h_{t}|h_{<t},r_{<t},r_t)}  \right]
\end{split}
\end{equation} 
which is analytical intractable to compute and usually optimized via the Monte Carlo methods.

Remarkably, the inference models in previous variational learning studies for sequential latent variable models \cite{kim2020sequential,zhang-etal-2020-probabilistic} are first-order Markov models, i.e. the latent state at current turn only depends on that at previous turn (e.g. $q_\phi(b_t|b_{t-1},r_{t-1},u_t,r_t)$ used in \cite{zhang-etal-2020-probabilistic}). In contrast, the session-level GPT-based inference model in VLS-GPT is inherently not a Markov model - the latent state at current turn $h_t$ depends on all history latent states $h_{1:t-1}$. The use of self-attention in the Transformer architecture connects current position to all previous positions.
The ELBO objective Eq. (\ref{eq:J_VL_2}) is thus an expectation under non-Markovian inference model. Its stochastic optimization presents new challenges, both \emph{statistically} and \emph{computationally}.
In the following, we first establish the Recursive Monte Carlo Approximation (RMCA) to the ELBO objective with non-Markovian inference model and prove its unbiasedness. 
Second, we develop the computational strategy of sampling-then-forward-computation to realize RMCA, which successfully overcomes the memory explosion issue of using GPT in variational learning and speeds up training.

\begin{algorithm}[t]
	\algsetup{linenosize=\small}
	\caption{Recursive Monte Carlo approximation with STT}
 	\label{alg:recursive}
	\begin{algorithmic}
	    \REQUIRE $u_{1:T},r_{1:T}$ with generative model $p_\theta$ in Eq. (\ref{eq:p_theta}), inference model $q_\phi$ in Eq. (\ref{eq:q_phi})
		\STATE $J=0$;
		\FOR{$t=1$ to $T$} 
		\STATE $i=1$;
		\STATE Given previous sampled states $h_{<t}$:
		\REPEAT
		\STATE Given previous sampled state tokens $h_t^{(<i)}$:
		\STATE $J += \sum_{\bar{h}_{t}^{(i)}} q_\phi(\bar{h}_{t}^{(i)}|STT(h_{<t}),r_{<t},r_t,STT(h_{t}^{(<i)}))$
		\STATE $~~~~~~~~~~~~~~~\times\log \frac{p_\theta(\bar{h}_{t}^{(i)}|STT(h_{<t}),r_{<t},STT(h_{t}^{(<i)}))}{q_\phi(\bar{h}_{t}^{(i)}|STT(h_{<t}),r_{<t},r_t,STT(h_{t}^{(<i)})}$;
		\STATE Draw $h_t^{(i)} \sim q_\phi(h_{t}^{(i)}|h_{<t},r_{<t},r_t,h_{t}^{(<i)})$;
        \STATE $i+=1$;
        \UNTIL {The $<eos>$ token is generated}
		\STATE $J += \log p_\theta(r_{t}|STT(h_{<t}),r_{<t},STT(h_t))$;
		\ENDFOR
		\STATE {\bf Return:} $J$
	\end{algorithmic}
\end{algorithm}

\subsection{Recursive Monte Carlo approximation to ELBO}
A naive Monte Carlo approximation is to draw one sample $h_{1:T} \sim q_\phi(h_{1:T}|u_{1:T},r_{1:T})$ and optimize the following estimator of the ELBO objective (via the STT trick):
\begin{displaymath}
\mathcal{J}_{\text{VL}} \approx
\sum_{t=1}^{T} \log p_\theta(r_{t}|h_{<t},r_{<t},h_t) +
\log \frac{p_\theta(h_{t}|h_{<t},r_{<t})}{q_\phi(h_{t}|h_{<t},r_{<t},r_t)}
\end{displaymath} 
This method is found to perform very unstable and fails to converge in our experiments, presumably due to the high variance of the Monte Carlo estimator.
Therefore, we propose the following recursive Monte Carlo approximation for VLS-GPT, as shown in Algorithm \ref{alg:recursive}, which has two main features. The first is to employ ancestral sampling according to the inference model, and the second is to calculate the KL divergences arised in the second term in the ELBO objective Eq. (\ref{eq:J_VL_2}) analytically as much as possible, so that the Monte Carlo variance is reduced \cite{kingma2013auto,kim2020sequential,zhang-etal-2020-probabilistic}.


Algorithm \ref{alg:recursive} summarizes the forward pass to calculate the ELBO objective with recursive Monte Carlo approximation.
Here follows several comments for illustration.
First, the latent state $h_t$ at any turn is a token sequence. Thus, the second term in the ELBO objective Eq.(\ref{eq:J_VL_2}), denoted by $\mathcal{J}_{\text{VL2}}$, can be further decomposed into a token-level sum:
\begin{equation} \label{eq:token-level-J}
\begin{split}
&\mathcal{J}_{\text{VL2}}\\
&=\mathbb{E}_{q_\phi(h_{1:T}|u_{1:T},r_{1:T})}\left[\sum_{t=1}^{T} \sum_{i=1}^{|h_t|} \log \frac{p_\theta(h_{t}^{(i)}|h_{<t},r_{<t},h_{t}^{(<i)})}{q_\phi(h_{t}^{(i)}|h_{<t},r_{<t},r_t,h_{t}^{(<i)}}  \right]
\end{split}
\end{equation}
The state tokens $h_t^{(i)}$ are recursively sampled until the special token $<eos>$ (end-of-sentence) is generated, and the length $|h_t|$ is thus determined.
At turn $t$ and position $i$, the expected log ratio between the prior and the posterior of current token, given previous sampled state tokens, turns out to be the KL divergence, which can be computed analytically.
Then, we sample $h_t^{(i)}$ and iterate to the next position.
After all the sampled tokens for turn $t$ are obtained, the first term in the ELBO objective Eq.(\ref{eq:J_VL_2}) can be directly estimated based on the sampled states.

Second, we show in Appendix \ref{ap:prop-1} that the following Proposition \ref{prop:recursive} holds, where we make explicit the dependence of $J$ on the sampled states $h_{1:T}$ and $\mathcal{J}_{\text{VL}}$ on $T$.
Proposition \ref{prop:recursive} is new and stronger in establishing the unbiasedness of such recursive Monte Carlo approximation to
the ELBO objective with non-Markovian inference model, beyond of those in \cite{kim2020sequential,zhang-etal-2020-probabilistic} which can be thought of as weak versions of RMCA, working with Markovian inference model.
\begin{prop} \label{prop:recursive}
The output $J(h_{1:T})$ from the recursive Monte Carlo approximation shown in Algorithm \ref{alg:recursive} is an unbiased estimator of the ELBO objective Eq.(\ref{eq:J_VL_2}), i.e.
\begin{equation} \label{eq:unbias}
\mathbb{E}_{q_\phi(h_{1:T}|u_{1:T},r_{1:T})} \left[ J(h_{1:T}) \right] = \mathcal{J}_{\text{VL}}(T) 
\end{equation}
\end{prop}

Third, taking the derivatives of $J(h_{1:T})$ w.r.t. $\theta$ and $\phi$ yields the stochastic gradients to update the model parameters.
Remarkably, Algorithm \ref{alg:recursive} not only shows the forward pass to obtain the stochastic estimator of the ELBO objective $J(h_{1:T})$, but also shows the application of the Straight-Through Trick (STT), as defined in Eq. (\ref{eq:ST-trick}), for calculating the gradients with discrete latent variables $h_t^{(i)}$'s.
The STT trick is applied to each sampled state tokens $h_t^{(i)}$'s in the forward pass for computing $J(h_{1:T})$.
Subsequently, in the backward pass, the gradients can be back-propagated through the sampled $h_t^{(i)}$'s for parameter update.



\begin{figure}[t]
	\centering
	\includegraphics[width=0.98\columnwidth]{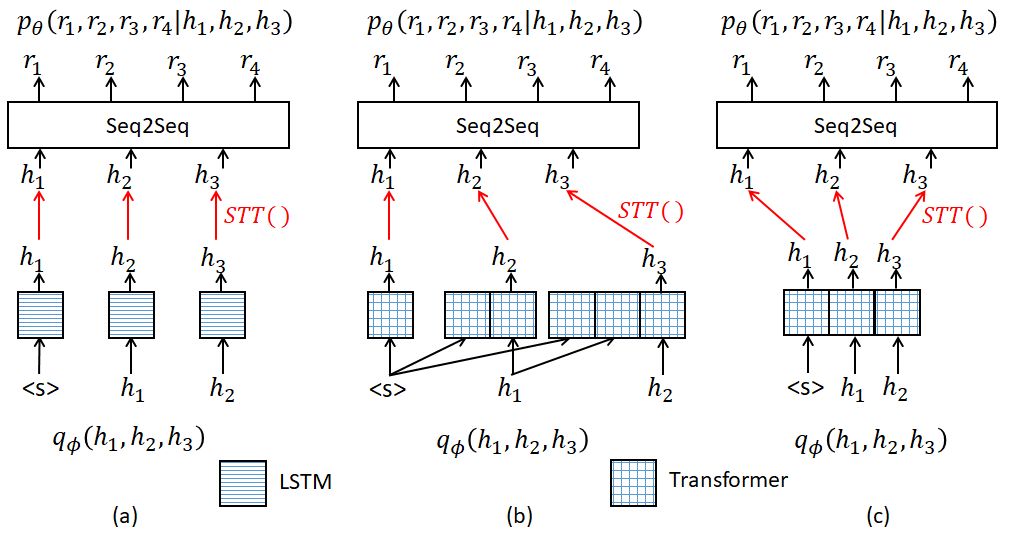}
	\caption{Illustration of forward calculation with different models for optimization in variational learning.
(a) $q_\phi(h_1,h_2,h_3)$ is a first-order Markov model. (b)(c) $q_\phi(h_1,h_2,h_3)$ is based on GPT, which is non-Markovian.
The difference between (b) and (c) is how the computational graph is created, which yields different memory costs. See text for details.
For (c), we run a forward pass first to infer $h_{1:T}$, which is omitted in the figure; only the second forward pass is shown here.
$STT()$ means applying Straight-Through Trick, as defined in Eq. (\ref{eq:ST-trick}).
}
	\label{fig:LSTM-vs-Transformer}
	\vspace{-1.0em}
\end{figure}

\subsection{Sampling-then-forward-computation strategy}
\label{sec:opt_stra}
As remarked above, the inference models in previous works \cite{kim2020sequential,zhang-etal-2020-probabilistic} are turn-level first-order Markovian. In contrast, the inference model in VLS-GPT is non-Markovian due to the use of the Transformer architecture.
The use of self-attention, connecting current position to all previous positions, leads to great memory consumption, if we apply the computational strategy as used in \cite{kim2020sequential,zhang-etal-2020-probabilistic} to realize RMCA to optimize the ELBO objective $J(h_{1:T})$.

For illustration shown in Figure \ref{fig:LSTM-vs-Transformer}\footnote{Without loss of generality, the illustration is taken at the turn level, without delving into the token level. In fact, the latent state $h_t$ at any turn is a token sequence. Thus, Figure \ref{fig:LSTM-vs-Transformer}(a), (b) and (c) should all be expanded by token-by-token sampling.}, we drop the conditional on $u_{1:T}$ and consider a simplified optimization $\max_{\theta,\phi}E_{q_\phi(h_1,h_2,h_3)} \left[ \log p_\theta(r_1,r_2,r_3|h_1,h_2,h_3) \right]$, which is similar to optimizing the actual ELBO objective function, namely optimizing an expectation under the inference model.
The computational strategy used in \cite{kim2020sequential,zhang-etal-2020-probabilistic} to realize RMCA is shown in Figure \ref{fig:LSTM-vs-Transformer}(a). In this strategy, turn-by-turn sampling of $h_{1:3}$ from $q_\phi(h_1,h_2,h_3)$ and feeding $h_{1:3}$ forward to compute $p_\theta(r_1,r_2,r_3|h_1,h_2,h_3)$ are taken in one forward pass, which creates the computational graph at the same time (with \emph{requires\_grad=true}).
This is feasible, since the model is turn-level first-order Markovian and the memory complexity of the computation graph is $O(T)$ ($T$ denotes the number of turns in a dialog).
If we apply this one-forward-pass strategy to realize RMCA for VLS-GPT, the memory complexity of the computation graph will be increased to $O(T(T+1)/2)$, as illustrated in Figure \ref{fig:LSTM-vs-Transformer}(b).

We propose a sampling-then-forward-computation strategy to realize RMCA for variational learning of VLS-GPT, as illustrated in Figure \ref{fig:LSTM-vs-Transformer}(c). We first run ancestral sampling of $h_{1:3}$ from $q_\phi(h_1,h_2,h_3)$ (with \emph{requires\_grad=false}), which is not shown in Figure \ref{fig:LSTM-vs-Transformer}. Then, we can treat the latent states $h_{1:3}$ as known, compute $q_\phi(h_1,h_2,h_3)$ in the forward direction, feed $h_{1:3}$ forward to compute $p_\theta(r_1,r_2,r_3|h_1,h_2,h_3)$ (with \emph{requires\_grad=true}). The resulting computational graph becomes much smaller, still in the complexity of $O(T)$.


Putting all together, applying the sampling-then-forward-computation strategy to realize RMCA to optimize the ELBO objective $J(h_{1:T})$ together with the Straight-Through trick, the unsupervised variational training of VLS-GPT is summarized as follows.
The semi-supervised training of VLS-GPT is shown in Algorithm \ref{alg:semi}.

\begin{algorithm}[t]
	\algsetup{linenosize=\small}
	\caption{Semi-supervised training of VLS-GPT}
 	\label{alg:semi}
	\begin{algorithmic}
	    \REQUIRE A mix of labeled and unlabeled dialogue data
	    \STATE Run supervised pre-training of $\theta$ and $\phi$ on labeled data;
	    \REPEAT
		\STATE Draw a labeled mini-batch of dialogs;
		\STATE Update $\theta$ and $\phi$ via supervised gradients;
        \STATE Draw an unlabeled mini-batch of dialogs;
        \FOR {an unlabeled dialog $u_{1:T},r_{1:T}$}
        \STATE Latent state generation (\emph{requires\_grad=false}): 
        \STATE ~~~~Draw $h_{1:T} \sim q_\phi(h_{1:T}|u_{1:T},r_{1:T})$; 
        \STATE Forward computation (\emph{requires\_grad=true}): 
        \STATE ~~~~Apply Algorithm \ref{alg:recursive}, but omit the step of sampling $h_{t}^{(i)}$'s, to obtain $J(h_{1:T})$;
        \STATE Backward computation and accumulate gradients;
        \ENDFOR
        \STATE Update $\theta$ and $\phi$ via unsupervised gradients;
        \UNTIL {convergence}
        \STATE {\bf Return:} $\theta$ and $\phi$
	\end{algorithmic}
\end{algorithm}

An iteration of unsupervised training consists of three steps - latent state generation, forward computation, backward computation.
First, we run sampling of $h_{1:T}$ (via greedy decoding in our experiments) from $q_\phi(h_{1:T}|u_{1:T},r_{1:T})$, which is termed as latent state generation.
Then in forward computation, we can apply Algorithm \ref{alg:recursive}, but treating $h_{1:T}$ as given, to obtain $J(h_{1:T})$.
Finally, we run the backward pass to obtain the gradients, which are used to update the generative model parameter $\theta$ and the inference model parameter $\phi$.



\section{Experiments}
	
	\subsection{Datasets} 
	We conduct our experiments on MultiWOZ2.1 \cite{eric2019multiwoz} and CrossWOZ \cite{zhu2020crosswoz}. MultiWOZ2.1 is a large-scale English multi-domain dialogue datasets of human-human conversations. Compared to MultiWOZ2.0, MultiWOZ2.1 removed noisy state values from the dialog state annotations. It contains 8438 multi-turn dialogues with 13.68 average turns, spanning over seven domains (restaurant, train, attraction, hotel, taxi, hospital, police) and providing additional validation set and test set, each of 1000 dialogues.
	
	CrossWOZ is the first large-scale Chinese Cross-Domain Wizard-of-Oz task-oriented dataset. It contains 6K dialogue sessions and 102K utterances for 5 domains, including hotel, restaurant, attraction, metro, and taxi. Moreover, the corpus contains rich annotation of dialogue states and dialogue acts at both user and system sides.
	
	\subsection{Data Pre-processing}
	\label{preprocess}
	We delexicalize dialog responses to reduce surface language variability on both datasets. During delexicalization, we replace values in the ontology with specific placeholders such as $[value\_name]$ and $[value\_price]$.
	We use the same pre-processing method as in UBAR \cite{yang2021ubar}, which implements domain-adaptive pre-processing like in DAMD \cite{zhang2020task}. This pre-processing method adopts a domain-adaptive delexicalization scheme, which decouples the domain and slot name of placeholders, by representing belief states as $[domain_1]~slot~value~slot~value~[domain_2]~slot~value$ sequences and representing system acts as $[domain]~[inform]~slot~[request]~slot$ sequences. The domains, acts and placeholders for slot values are all bracketed as special tokens.
	\revfirst{
	Remarkably, to interact with real users, the system will lexicalize the generated  delexicalized responses using the generated belief states and the entities queried from the database, which currently is a common practice.}
	
	\subsection{Metrics} \label{sec:metric}
	In our experiments on MultiWOZ2.1, we follow the original MultiWOZ guidance \cite{budzianowski2018large} for individual metrics and follow \cite{mehri2019structured} for the combined score. \emph{Inform Rate} measures how often the entities provided by the system are correct. \emph{Success Rate} refers to how often the system is able to answer all the requested attributes by user. \emph{BLEU Score} is used to measure the fluency of the generated responses by analyzing the amount of n-gram overlap between the real responses and the generated responses. And \emph{Combined Score} is computed as (BLEU + 0.5 * (Inform + Success)). 
	
	
	As for CrossWOZ, we develop end-to-end corpus-based evaluation scripts, which are missing in the original release of CrossWOZ. 
	In MultiWOZ, the Inform and Success metrics are computed in session-levels, which means entity matching and success can only be 0 or 1 for a dialog.
We propose to use finer grained metrics on CrossWOZ, considering its characteristics. 
\emph{Match rate} is a turn-level metric to measure the system's ability to provide correct entities, which is obtained by calculating the proportion of turns providing correct entities in all turns that provide entities. 
\emph{Request Success rate} (Req-Suc) is also a turn-level metric, namely the proportion of informative attributes in oracle system acts that appear in generated responses, which reflects the system's ability to successfully answer user requests.
\emph{BLEU} measures the fluency of generated responses. 
\emph{Combined Score} is computed as (BLEU + 0.5 * (Match + Req-Suc)).

Note that different from MultiWOZ, users in CrossWOZ may ask for multiple entities with different constraints in the same domain at different turns. For example, the user wants to eat in both a roast duck restaurant and a pancake restaurant. The user asked about the two types of restaurants in two different turns and the system must provide correct entities respectively. It is better to calculate Match rate turn by turn in this case.
Req-Suc does not check the matching of entities again, since turn-level entity matching is already evaluated by Match rate.
	
	
	\subsection{Implementation Details}
\label{sec:train_detail}
All models are trained on two 16-GB Tesla P100 GPUs. The training time of one semi-supervised experiment (Semi-ST or Semi-VL) with a certain label proportion in Table \ref{semi-results} is about two days. 
We implement the models with Huggingface Transformers repository of version 3.5.1. We initialize the generative and inference models with DistilGPT2 which is a distilled version of GPT-2 and has 6 self-attention layers. The maximum sequence length is 1024 and sequences that exceed 1024 tokens are pre-truncated. We use the AdamW optimizer and a linear scheduler with 20\% warm-up steps. We run 50 epochs during supervised pre-training and 40 epochs during semi-supervised learning. Early stopping is not used in our experiment and we select the model of the highest combined score on validation set during training. The maximum learning rate of linear scheduler is 1e-4 and the batch size is 32 dialogs, which is implemented with basic batch size of 2 and gradient accumulation steps of 16. During evaluation, we use the greedy decoding method and generate latent states and responses in batches with the past-key-values mechanism to reduce time consuming.
We will release the code when this work is published.

	\subsection{Fully-supervised Baselines}
	\label{baseline}
	In this section, we show the results of end-to-end modeling and evaluation in the fully-supervised setting, where the models, trained with 100\% labeled data, are used to generate belief states, query database with the generated
	belief states, and then generate acts and responses. 
	\revfirst{
	In the fully-supervised setting, only the generative model in \modelname{}, namely \modelname{}-p, is trained and tested.}
	We compare \modelname{}-p with other task-oriented end-to-end models including LABES \cite{zhang-etal-2020-probabilistic}, SimpleTOD \cite{hosseini2020simple}, AuGPT \cite{kulhanek2021augpt} and UBAR \cite{yang2021ubar}.
	The main purpose of the fully-supervised experiments is to gauge the strength of the generative model \modelname{}-p.
	The results are shown in Table \ref{sup-results}. 
	\begin{table}[t]
		\caption{End-to-end evaluation results on fully-supervised MultiWOZ2.1. * denotes results obtained by our run of the open-source code. 
		The means and standard deviations for UBAR and \modelname{}-p are from 3 independent runs. 
		}
		\centering
		\resizebox{\linewidth}{!}{
			\begin{tabular}{cc cccc}
				\toprule
				Model &Pretrained LM &Inform &Success &BLEU &Combined\\
				\midrule
				DAMD &-  &76.4 &60.4 &16.6 &85.0\\
				LABES-S2S &-  &76.89 &63.3 &17.92 &88.01\\
				SimpleTOD &DistilGPT-2  &85.00 &70.05 &15.23 &92.98\\
				AuGPT &GPT-2  &91.4 &72.9 &17.2 &99.35\\
         UBAR$^*$ &DistilGPT-2 &89.62$\pm${0.56} &80.85$\pm${1.03} &17.60$\pm${0.13} &102.84$\pm${0.38}\\
         \revfirst{\modelname{}-p} &DistilGPT-2 &90.27$\pm${0.53} &81.44$\pm${0.82} &17.48$\pm${0.16} &103.33$\pm${0.91}\\
         \bottomrule	
		\end{tabular}}
		\label{sup-results}
	\end{table}

	Table \ref{sup-results} shows that \modelname{}-p obtains state-of-the-art results on MultiWOZ2.1, compared to other recent models in an end-to-end evaluation\footnote{Note that the end-to-end results reported in UBAR's original paper \cite{yang2021ubar} are obtained through an incomplete end-to-end evaluation, where the oracle belief states are used for database query. When also using this trick in our evaluation, \modelname{}-p obtains a combined score of 106.6, which is higher than 105.7 reported in UBAR.}.
	Considering that the generative model \modelname{}-p is similar to UBAR (but can be suited to variational learning), and their results are close to each other, the two models were run with 3 random seeds.
	Further, taking each testing dialog as a sample, we conduct the matched-pairs significance test \cite{gillick1989some} to compare fully-supervised VLS-GPT-p and UBAR.
The p-values for Inform, Success and BLEU are 0.42, 0.87, 0.23, respectively.
Overall, these results show that fully-supervised VLS-GPT-p achieves minor improvement over as UBAR (not significantly better). Enhancing the fully-supervised baseline is not the main focus of this paper.


	\begin{table*}[t]
	    \caption{Semi-supervised results on MultiWOZ2.1 and CrossWOZ. \revfirst{All results are reported as the means from 3 independent runs with different random seeds. The standard deviations are shown by the error bars in Figure \ref{fig:score-proportions}.}}
		\centering
		\resizebox{0.95\linewidth}{!}{
				\begin{tabular}{cc cccc cccc}
					\toprule
					\multicolumn{2}{c}{Model Configuration} &\multicolumn{4}{c}{MultiWOZ2.1} &\multicolumn{4}{c}{CrossWOZ}\\
					\cmidrule(lr){1-2} \cmidrule(lr){3-6} \cmidrule(lr){7-10}
					Proportion &Method  &Inform &Success &BLEU &Combined  &Match &Req-Suc &BLEU &Combined\\
					\midrule
					100\% &SupOnly  &90.27 &81.44 &17.48 &103.33 &61.88 &75.77 &33.81 &102.63\\
					\midrule
					\multirow{3}{*}{50\%} &SupOnly   &82.95 &72.37 &16.74 &94.40 &60.68 &73.03 &27.95 &94.81\\
					&Semi-ST  &84.95 &72.44 &16.54 &95.24 &61.28 &73.15 &29.66 &96.87\\
					&Semi-VL  &87.39 &77.61 &16.71 &99.21 &60.65 &72.71 &29.54 &96.22\\
					\midrule
					\multirow{3}{*}{40\%} &SupOnly   &82.35 &70.70 &16.43 &92.95 &60.01 &73.69 &26.80 &93.65\\
					&Semi-ST  &81.31 &69.60 &16.18 &91.64 &60.29 &70.11 &28.89 &94.09\\
					&Semi-VL  &84.68 &72.64 &16.46 &95.12 &61.25 &74.61 &29.80 &97.74\\
					\midrule
					\multirow{3}{*}{30\%} &SupOnly &77.78 &66.37 &15.81 &87.89 &58.62 &71.48 &25.92 &90.98 \\
					&Semi-ST  &77.68 &66.67 &16.22 &88.39 &59.73 &71.07 &29.82 &95.23\\
					&Semi-VL  &84.89 &74.17 &16.59 &96.12 &59.44 &73.83 &28.93 &95.56\\
					\midrule
					\multirow{3}{*}{20\%} &SupOnly  &71.34 &58.06 &15.33 &80.03 &58.56 &67.46 &24.49 &87.50\\
					&Semi-ST  &73.61 &62.39 &15.61 &83.61 &57.46 &68.82 &27.38 &90.52\\
					&Semi-VL  &79.41 &68.54 &16.54 &90.52 &59.37 &71.67 &29.93 &95.45\\
					\midrule
					\multirow{3}{*}{10\%} &SupOnly  &56.59 &42.14 &13.40 &62.77 &53.48 &67.62 &20.41 &80.96\\
					&Semi-ST  &71.94 &57.96 &15.20 &80.15 &54.78 &67.59 &24.19 &85.37\\
					&Semi-VL  &76.58 &65.63 &15.01 &86.12 &58.38 &71.00 &27.97 &92.66\\
					\bottomrule
			\end{tabular}}
			\label{semi-results}
	\end{table*}

\subsection{Semi-Supervised Experiments}
	\label{semi-sup-exp}
	\revfirst{Some proportions of the labeled dialogs from MultiWOZ2.1 training set are randomly drawn, with the rest dialogs treated as unlabeled.
	In supervised-only training, denoted by SupOnly, the rest dialogs are discarded and only the generative model VLS-GPT-p is trained.
	Different semi-supervised models are trained in two stages. 
	The first stage is supervised pre-training of VLS-GPT-p and VLS-GPT-q (if used) over labeled data only.
	The second stage is semi-supervised learning over both labeled and unlabeled data.
	Semi-supervised models could be implemented by the variational learning method (Semi-VL) or the self-training (Semi-ST) baseline method.
	Semi-VL stands for exactly what VLS-GPT does, as shown in Algorithm \ref{alg:semi}.
	Self-training (ST), also known as pseudo-labeling, is a classic strong semi-supervised learning method. It uses only the generative model \modelname{}-p and performs as its name suggests, i.e. generating hypothesized labels using the current model and then perform supervised training with the pseudo-labeled samples to update the model.
	See Section \ref{sec:self-training} for more details about ST.}
	
\begin{figure*}
	\centering
	\subfigure[]{
		\includegraphics[width=0.49\linewidth]{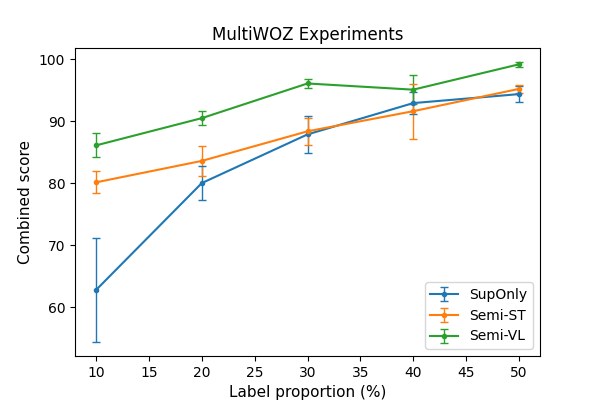}}
	\subfigure[]{
	    \includegraphics[width=0.49\linewidth]{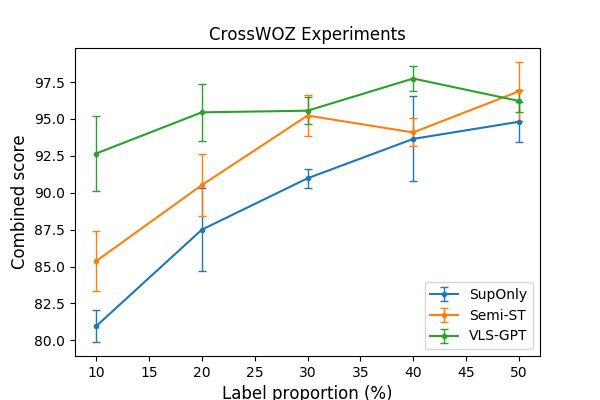}}
		\caption{Combined Scores at different label proportions on MultiWOZ2.1 and CrossWOZ. The standard deviations are shown by the error bars.}
		\vspace{-0.8em}
		\label{fig:score-proportions}
	\end{figure*}

	We conduct semi-supervised experiments with different labeling proportions from 10\% to 50\%. The results on MultiWOZ2.1 and CrossWOZ are shown in Table \ref{semi-results}. The combined scores against label proportions with standard deviations are shown in Figure \ref{fig:score-proportions}.
	The main observations are as follows.
	
	First, we can see that the two semi-supervised methods (Semi-ST and Semi-VL) generally outperform the SupOnly method across the two datasets of different languages and at different label proportions. 
	This clearly demonstrate the advantage of semi-supervised TOD systems.
	A few results where Semi-ST performs worse than SupOnly may reflect some instability of Semi-ST.
	
	Second, when comparing the two semi-supervised methods, Semi-VL generally performs better than Semi-ST across different languages and label proportions. 
	A close look at Table \ref{semi-results} reveals that the improvements of Semi-VL over Semi-ST are much larger in Match Rate and Success Rate than in BLEU.
	Remarkably, the Inform and Success metrics depend on the capability of a particular method for predicting hidden states (belief states and system acts). In contrast, BLEU measures the fluency of generated responses and may be improved just by observing more (unlabeled) responses.
	In semi-supervised experiments, the system responses are observed in both methods of Semi-VL and Semi-ST, which may make BLEU results across different methods differ not much.
Therefore with the above analysis, better Inform and Success of Semi-VL than Semi-ST indicate the superiority of Semi-VL in learning from unlabeled data to improve the prediction accuracy of belief states and system acts, not merely to improve BLEU.
   
   \revfirst{Third, From Table \ref{semi-results}, careful readers may find that Semi-VL outperforms Semi-ST with a large margin on MultiWOZ 2.1, while it only slightly outperforms Semi-ST on CrossWOZ. 
   Presumably, this difference is caused by the more complexity of CrossWOZ, compared to MultiWOZ.
   The average number of mentioned domains per dialog in CrossWOZ is 3.24, while it is 1.80 in MultiWOZ. 
   Moreover, users in CrossWOZ may ask for multiple entities with different constraints in the
same domain at different turns, as introduced in Section\ref{sec:metric}; and there are many co-references when users query nearby entities.
The more complexity of the dialog tasks in CrossWOZ increases the difficulty for both Semi-VL and Semi-ST in predicting belief states in many cases.
As long as the predicted belief states are not completely correct, the results produced by both methods will be counted as failures and the difference between the metrics from the two methods will become smaller.}

	Fourth, notably, combining Table \ref{sup-results} and \ref{semi-results}, we can see that Semi-VL of VLS-GPT with only 20\% labeled data already performs better than fully-supervised LABES (namely with 100\% labeled data).
	Moreover, it is observed that Semi-VL of VLS-GPT with 50\% labeled data performs close to the fully-supervised VLS-GPT.
	These results clearly show that the benefit of combining the strengths of both pre-trained GPT and variational learning for semi-supervisded TOD systems.
	Dialog examples are provided in Section \ref{sec:case} to understand the superiority of Semi-VL over SupOnly and Semi-ST.
	
	Finally, from the plot of metric scores against labeling proportions in Figure \ref{fig:score-proportions}, we observe that the smaller proportion of labels, the larger gain obtained by the semi-supervised methods. The semi-supervised methods can significantly improve the performance when the label proportion is as small as 10\%, which demonstrates the fast learning capability of the semi-supervised learning methods.

\revsecond{
   \subsection{The performance of inference model}
As suggested by a referee, we examine the performance of the inference models in inferring the latent states (belief states, DB results and system actions). We consider the two inference models, which are obtained by the two methods of SupOnly and Semi-VL respectively with 10\% labeled data on MultiWOZ2.1. The ground truth latent states and the inferred latent states (via greedy decoding with the inference models) are compared on the test set of MultiWOZ 2.1. The joint goal accuracy (Joint Goal) and slot F1 score (Slot F1) for belief states, DB result accuracy (DB acc), and system act F1 score (Act F1) are calculated and the results are shown in Table~\ref{tab:infer-perform}.
It can be seen that the Joint Goal, Slot F1 and DB acc of the inference model of Semi-VL are substantially increased, when compared to the inference model of SupOnly. This shows the advantage of variational learning. On the other hand, it is interesting to see that the Act F1 becomes worse after Semi-VL.
Notably, the variational ELBO objective Eq. (\ref{eq:J_VL_2}) consists of two terms, and the second term is to minimize the KL divergence of the approximate posterior from the prior of latent states, which acts as a regularizer \cite{kingma2013auto}. 
Note that the prior for $a_t$ (i.e., determining $a_t$ from $u_t$ without knowing $r_t$) is dramatically different from its posterior (i.e., determining $a_t$ with both $u_t$ and $r_t$), while less so for $b_t$ (i.e., determining $b_t$ from $u_t$ with $r_t$ or not).
Thus, pushing the posterior closer to the prior will presumably have more adverse effect on learning the posterior of $a_t$ than on that of $b_t$. This reveals some shortcoming of variational learning and points to interesting future work.}
   \begin{table}[t]
		\caption{\revsecond{The performance of the inference models trained by different methods with 10\% labeled data on MultiWOZ2.1.}}
		\centering
		\resizebox{0.95\linewidth}{!}{
				\begin{tabular}{cc cccc}
					\toprule
					\multicolumn{2}{c}{Model Configuration} &\multicolumn{4}{c}{Metrics} \\
					\cmidrule(lr){1-2} \cmidrule(lr){3-6}
					Proportion &Method  &Joint Goal &Slot F1 &DB acc &Act F1\\
					\midrule
					10\% & SupOnly & 28.01 &77.37 & 76.03 & 82.87\\
                        10\% & Semi-VL & 35.94 &84.81 & 83.97 & 76.16\\
					\bottomrule
			\end{tabular}}
		\vspace{-0.5em}
		\label{tab:infer-perform}
	\end{table}

	\subsection{Complexity analysis} \label{sec:complexity}
	Recall that an iteration in Semi-VL consists of three steps - latent state generation, forward computation, backward computation.
	Due to its auto-regressive nature, the generation process of GPT-2 is very slow and latent state generation in Semi-VL consumes large amounts of training time.
	Take running Semi-VL with 20\% labels on MultiWOZ2.1 in two 16-GB Tesla P100 GPUs as an example. The three steps for an epoch take 32 minutes, 12 minutes and 12 minutes respectively.
	In the proposed sampling-then-forward-computation strategy, we first use the inference model to generate latent states without gradients (\emph{requires\_grad=false}), so that we can use a much larger batch size of 32 in latent state generation. 
	In contrast, if we use the previous strategy of coupling sampling and forward computation in one pass, the affordable batch size is 2 and such one-forward-pass takes 300 minutes for an epoch.
	Thus, the proposed strategy achieves a speedup by 7-fold (300/(32+12)). 
	In summary, the proposed strategy of sampling-then-forward-computation in training not only reduces the memory cost, but also accelerates latent state generation substantially. 

	\begin{table}[t]
	    \caption{Ablation experiments on different semi-supervised self-training schemes with 10\% labeled data on MultiWOZ2.1.
		}
		\centering
		\resizebox{\linewidth}{!}{
			\begin{tabular}{l ccccc}
				\toprule
				Scheme  &Inform &Success &BLEU &Combined\\
				\midrule
				$\mathcal{J}_{\text{ST-response}} \text{~with~} STT(h_t^{(i)})$ &71.94 &57.96 &15.20 &80.15\\
				$\mathcal{J}_{\text{ST-joint}} \text{~with~} STT(h_t^{(i)})$ &58.86 &49.85  &14.75  &69.10\\
				$\mathcal{J}_{\text{ST-response}}$ &68.02 &54.45 &14.93 &76.17\\
				$\mathcal{J}_{\text{ST-joint}}$ &47.35 &38.24  &13.80  &56.59\\
				\revfirst{ST with inference model} &66.56 &51.85 &12.54 &71.75\\
				\bottomrule
		\end{tabular}}
		\label{semi-ST-ablation}
	\end{table}

	\subsection{On the self-training semi-supervised method} \label{sec:self-training}
	Notably, applying self-training to the generative model \modelname{}-p is different from applying self-training to an ordinary classifier, and there are several possible schemes. This section introduces more experiments on the self-training methods, and we choose the strongest among possible self-training schemes as the Semi-ST method, which is reported in Table \ref{semi-ST-ablation} to compare with Semi-VL.
	
	In self-training, given unlabeled dialog $\{u,r\}_{1:T}$, we generate hypothesized label $h_{1:T}$ via greedy decoding based on the latent state prior $\sum_{t=1}^{T} \log p_\theta(h_t| h_{<t},r_{<t})$, and then use the pseudo-labeled $h_{1:T}$ to update the generative model parameter $\theta$ by maximizing the response probability,
	\begin{displaymath}
	\mathcal{J}_{\text{ST-response}}=\sum_{t=1}^{T} \log p_\theta(r_t|h_{<t},r_{<t},h_t)
	\end{displaymath}
	or the joint probability
	\begin{displaymath}
	\mathcal{J}_{\text{ST-joint}}=\sum_{t=1}^{T} \left[ \log p_\theta(h_t| h_{<t},r_{<t})+\log p_\theta(r_t|h_{<t},r_{<t},h_t)\right].
	\end{displaymath}
	In forward calculation of either objective function, we can apply $STT(h_t^{(i)})$ and thus the gradients will propagate through the discrete $h_{1:T}$, while classic self-training does not use STT.
	
	\revfirst{Notably, self-training typically involves only one model, i.e. \modelname{}-p here. The model used for prediction in testing is used for predicting pseudo labels in training.
	As suggested by a referee, we experiment with a variant of self-training, which uses not only \modelname{}-p but also \modelname{}-q. This ST scheme involves two models and is referred to as ``ST with inference model''.
	Specifically, we first use labeled data to train \modelname{}-q, which is then used to predict pseudo labels for unlabeled data. Finally, both labeled data and pseudo-labeled data are used to train the generative model \modelname{}-p in a supervised manner.}
    
	Table \ref{semi-ST-ablation} shows the semi-supervised results for the \revfirst{five} possible schemes of self-training with 10\% labeled data on MultiWOZ2.1.
	It can be seen that using $\mathcal{J}_{\text{ST-response}}$ with Straight-Through performs the best, which is exactly the Semi-ST used in Table \ref{semi-results} for comparing with Semi-VL and represents a strong semi-supervised baseline.
	
 	\revfirst{
	Presumably, the performance superiority of Semi-VL over the self-training methods comes from introducing the inference model for hypothesis generation and optimizingbased on the solid variational learning principle. 
	The first four ST only use the prior $p_\theta(h_t| h_{<t},r_{<t})$ for hypothesis generation.
	In contrast, Semi-VL uses the inference model via the posterior $q_\phi(h_t|h_{<t},r_{<t},r_t)$, and thus can exploit more information from $r_t$ to infer belief states and system acts. 
	Remarkably, the performance of  ``ST with inference model'' is moderate among the ST schemes.
	It seems that simply introducing an inference model, through supervised pre-training, to predict pseudo labels is inferior to Semi-VL. Importantly, the inference model in Semi-VL is optimized based on the solid variational learning principle. 
	This is beneficial for the inference model in Semi-VL to learn to generate better pseudo-labeled samples.
 }

	\begin{table}[t]
    \caption{\revfirst{Data augmentation (back-translation) results on MultiWOZ2.1.
    All results are reported as the means from 3 independent runs with different random seeds and thus are comparable to results shown in Table~\ref{semi-results}.}}
		\centering
		\resizebox{\linewidth}{!}{
			\begin{tabular}{ccccc}
				\toprule
				Proportion  &Inform &Success &BLEU &Combined\\
				\midrule
				50\% &83.15 &72.21 &16.85 &94.53\\
				40\% &81.78 &70.30 &16.47 &92.51\\
				30\% &78.31 &67.67 &15.90 &88.89\\
				20\% &70.27 &59.26 &15.21 &79.97\\
				10\% &54.02 &41.47 &12.86 &60.61\\
				\bottomrule
		\end{tabular}}
		\label{tab:augmentation}
	\end{table}
	
    \revfirst{\subsection{Comparison with data augmentation}
In addition to semi-supervised learning, a widely-used method to improve system performance in low resource scenarios is data augmentation. 
Data augmentation (DA) is a technique that augments the labeled training set with label-preserving synthetic samples. 
An effective DA method for TOD systems is paraphrasing via back-translation, as shown in AuGPT \cite{kulhanek2021augpt}.
In AuGPT, a trained multilingual machine translation model \cite{machavcek2020elitr} is employed with ten intermediate languages, and a set of different paraphrases for each input utterance is obtained.
We use the paraphrased data released by AuGPT at GitHub\footnote{\url{https://github.com/ufal/augpt/}} and conduct experiments in the same low resource settings as in Table~\ref{semi-results}.
Specifically, some proportions of the labeled dialogs from the MultiWOZ2.1 training set are drawn and paraphrased, which are used to train the generative model VLS-GPT-p.
In training, as in AuGPT, we choose the input user utterance uniformly at random from the set of all variants of the utterance including the original one.
    The results are shown in Table~\ref{tab:augmentation}. We can see that the models trained with augmented data perform slightly better than the SupOnly baseline in Table~\ref{semi-results} at the labeling proportions of 50\% and 30\%, while they are inferior to SupOnly at other proportions. The proposed Semi-VL outperforms the back-translation DA method at all proportions significantly. Presumably, such performance difference may be attributed to the fact that semi-VL can exploit not only the labeled data but also the unlabeled data, while back-translation only augments the labeled data.}
    
\subsection{Case Study}
\label{sec:case}
We provide \revfirst{a lexicalized testing example} in MultiWOZ2.1 in Table \ref{case}.
It can be seen that the supervised-only (SupOnly) baseline fails to predict the correct belief state, while Semi-VL makes the correct prediction.
The SupOnly model misses the generation of the $pricerange$ slot and its corresponding value $expensive$. Due to the incorrect belief state, the SupOnly model gets the wrong database result, and generates a completely inappropriate response. The belief state generated by the Semi-ST model contains some error but does not affect the generated response. 
The Semi-VL model outperforms the previous two models and generates both belief state and response perfectly. 

\begin{table}[t]

	\centering
	\caption{\revfirst{A lexicalized testing example} in MultiWOZ2.1. The belief states are represented in the form of $[domain]~ slot~ value~ slot~ value$. The original ground truth dialogs are in the top row. The label proportion of SupOnly, Semi-ST and Semi-VL models is 20\%.}
	\resizebox{.99\columnwidth}{!}{ 
		\begin{tabular}{cccc}
			\toprule
			\multicolumn{4}{c}{Dialog mul1211 in MultiWOZ2.1} \\
			\midrule
			\multicolumn{4}{l}{$u_1$: I would like to find in Cambridge, a restaurant, located in the} \\
			\multicolumn{4}{l}{~~~~~  centre area. I love Spanish food.}\\
			\multicolumn{4}{l}{$b_1$: [restaurant] food Spanish area centre.} \\
			\multicolumn{4}{l}{$r_1$: I have La Tasca downtown here at 14 - 16 Bridge street.} \\
			\specialrule{0em}{2pt}{2pt}
			\multicolumn{4}{l}{$u_2$: The restaurant should be in the expensive price range.}\\
			\multicolumn{4}{l}{$b_2$: [restaurant] food Spanish area centre pricerange expensive}\\
			\multicolumn{4}{l}{$r_2$: I do not have any in the centre, another criteria perhaps?}\\
			\midrule
			\multirow{4}{*}{SupOnly} & \multicolumn{3}{l}{$b_2$: [restaurant] food Spanish area centre.}\\
			& \multicolumn{3}{l}{$r_2$: La Tasca serves Spanish food and is in the}\\
			& \multicolumn{3}{l}{~~~~~centre area. It is in the moderate pricerange}\\
			& \multicolumn{3}{l}{~~~~~and the La Tasca serves Spanish food.}\\
			\midrule
			\multirow{5}{*}{Semi-ST} & \multicolumn{3}{l}{$b_2$: [restaurant] food Spanish area centre name the}\\
			& \multicolumn{3}{l}{~~~~~ pricerange expensive.}\\
			& \multicolumn{3}{l}{$r_2$: I am sorry, there are no expensive Spanish}\\
			& \multicolumn{3}{l}{~~~~~restaurants in the centre. Would you like to}\\
			& \multicolumn{3}{l}{~~~~~try a different type of food?}\\
			\midrule
			\multirow{5}{*}{Semi-VL} & \multicolumn{3}{l}{$b_2$: [restaurant] food Spanish area centre pricerange}\\
			& \multicolumn{3}{l}{~~~~~ expensive.}\\
			& \multicolumn{3}{l}{$r_2$: I am sorry, there are no expensive Spanish}\\
			& \multicolumn{3}{l}{~~~~~restaurants in the centre. Would you like to}\\
			& \multicolumn{3}{l}{~~~~~try another area  or pricerange?}\\
			\bottomrule
	\end{tabular}}
	\label{case} 
\end{table}
Table \ref{case1} shows an example in MultiWOZ2.1, which helps to illustrate why unlabled dialog data are helpful in learning TOD systems.
Intuitively, there are cues from user inputs and system responses, which reveal the belief states, database results and system acts. So the dialog data, even unlabeled, can be used to enhance the performance of belief tracking and action selection, and thus benefit the whole dialog system. 
\begin{table}[t]
	\centering
	\caption{An example in MultiWOZ2.1. The cues for belief states, database results and system acts are contained in user inputs and system responses, and are marked in red, blue and green respectively. The query result [db\_x] (x=0,1,2,3) indicates 0, 1, 2-3, and \textgreater3 matched results respectively.}
	\resizebox{.99\columnwidth}{!}{ 
		\begin{tabular}{l}
			\toprule
			Dialog sng0601 in MultiWOZ2.1\\
			\midrule
			$u_1$: I would like to go to an \textcolor{red}{Indian} restaurant in the \textcolor{red}{north}. \\
			$b_1$: [restaurant] food indian area north\\
			$d_1$: [db\_2]\\
			$a_2$: [restaurant] [select] price [inform] choice\\
			$r_1$: I found \textcolor{blue}{2} that matches your criteria. Would you prefer a \\
			~~~~~~~moderate or cheap pricing?\\
			\specialrule{0em}{2pt}{2pt}
			$u_2$: How about the \textcolor{red}{moderate} one? May I have their address, please?\\
			$b_2$: [restaurant] food indian area north pricerange moderate\\
			$d_2$: [db\_1]\\
			$a_2$: [restaurant] [inform] address name postcode [general] [reqmore]\\
			$r_2$: Yes the Nirala's \textcolor{green}{address} is 7 milton road chesterton and their\\
			~~~~~~~\textcolor{green}{postcode} is cb41uy. Is there anything else i can help you with\\
			~~~~~~~today?\\
			\specialrule{0em}{2pt}{2pt}
			$u_3$: No, that is all, thank you. Have a nice day.\\
			$b_3$: [restaurant] food indian area north pricerange moderate \\
			~~~~~~~name the nirala\\
			$d_3$: [db\_1]\\
			$a_3$: [general] [bye]\\
			$r_3$: So glad we could help you out. Thanks for using the cambridge\\
			~~~~~~ towninfo centre, and have a glorious day!\\
			\bottomrule
	\end{tabular}}
	\label{case1} 
\end{table}

Table \ref{tab:crosswoz} is an example from CrossWOZ testing set. The user utterance informs the constraint of ``duration'' and requests about the fee and surrounding restaurants. Among the three models, only Semi-VL generates the correct belief state. SupOnly generates ``fee 1\_hour'', which is false. Semi-ST mistakenly adds a slot-value pair ``fee free''.

\section{Conclusion and Future Work}
In this paper, we propose Variational Latent-State GPT model (VLS-GPT), which, to the best of our knowledge, is the first to combine the strengths of large pre-trained language model and variational learning for semi-supervisded TOD systems.
Due to the use of the Transformer architecture, the inference model in VLS-GPT is non-Markovian.
The variational ELBO objective is thus an expectation under non-Markovian inference model. Its stochastic optimization presents new challenges, both statistically and computationally, compared to previous variational learning works for sequential latent variable models, which use turn-level first-order Markovian.
In this work, we establish Recursive Monte Carlo Approximation (RMCA) to ELBO with non-Markovian inference model and prove its unbiasedness. Further, we develop the computational strategy of sampling-then-forward-computation to realize RMCA, which successfully overcomes the memory explosion issue of using GPT in variational learning and speeds up training.

Semi-supervised TOD experiments are conducted on two benchmark multi-domain datasets - MultiWOZ2.1 in English and CrossWOZ in Chinese. VLS-GPT is shown to outperform the supervised-only baseline, the strong semi-supervised GPT-based self-training baseline, and the variational learning only baseline, across languages.

\revfirst{
Remarkably, the recursive Monte Carlo approximation to ELBO with non-Markovian inference model and the computational strategy of sampling-then-forward-computation are useful in general for variational training of Transformer based latent variable models.
On top of VLS-GPT, there are interesting directions for future work.
First, it is interesting to extend VLS-GPT to leverage unlabeled open-domain data together with in-domain data for better semi-supervised learning of TOD systems.
Second, as overviewed in Section \ref{sec:variational-latent}, 
variational latent variable models can be used in TOD systems to enhance not only semi-supervised learning but also reinforcement learning.
While this paper mainly develops GPT based variational latent variable models for semi-supervised learning of TOD systems, it is definitely worthwhile to investigate the utilization of the RMCA and the sampling-then-forward-computation methods to learn GPT based latent action models for reinforcement learning of TOD systems. Hopefully this may be realized by marrying LaRL or LAVA-type models with some variant of VLS-GPT.
}



\appendices
\section{Proof of Proposition \ref{prop:recursive}} \label{ap:prop-1}

\begin{algorithm}[t]
	\algsetup{linenosize=\small}
	\caption{Turn-level Recursive Monte Carlo approximation}
 	\label{alg:recursive-turn}
	\begin{algorithmic}
	    \REQUIRE $u_{1:T},r_{1:T}$ with generative model $p_\theta$ in Eq. (\ref{eq:p_theta}), inference model $q_\phi$ in Eq. (\ref{eq:q_phi})
		\STATE $F=0$;
		\FOR{$t=1$ to $T$} 
		\STATE Given previous sampled states $h_{<t}$:
		\STATE $F += \sum_{\bar{h}_{t}} q_\phi(\bar{h}_{t}|h_{<t},r_{<t},r_t) \log \frac{p_\theta(\bar{h}_{t}|h_{<t},r_{<t})}{q_\phi(\bar{h}_{t}|h_{<t},r_{<t},r_t)}$
		\STATE Draw $h_t \sim q_\phi(h_{t}|h_{<t},r_{<t},r_t)$;
		\STATE $F += \log p_\theta(r_{t}|h_{<t},r_{<t},h_t)$;
		\ENDFOR
		\STATE {\bf Return:} $F$
	\end{algorithmic}
\end{algorithm}

\begin{proof}
Note that for both the generative model $p_\theta$ in Eq. (\ref{eq:p_theta}) and the inference model $q_\phi$ in Eq. (\ref{eq:q_phi}), the auto-regressive structures at the token-level are very close to those at the turn-level. This analogy can also be seen from the similarity between the token-level sum in Eq. (\ref{eq:token-level-J}) and the turn-level sum in Eq.(\ref{eq:J_VL_2}). 
Thus, without loss of generality, we mainly prove the unbiasedness of the turn-level recursive Monte Carlo approximation shown in Algorithm \ref{alg:recursive-turn}, i.e.
\begin{equation} \label{eq:unbias-2}
\mathbb{E}_{q_\phi(h_{1:T}|u_{1:T},r_{1:T})} \left[ F(h_{1:T}) \right] = \mathcal{J}_{\text{VL}}(T) 
\end{equation}
The unbiasedness of the token-level recursive Monte Carlo approximation shown in Algorithm \ref{alg:recursive} can be proved analogously.

First, Eq. (\ref{eq:unbias-2}) clearly holds for $T=1$. Then, we proceed by mathematical induction.
Suppose Eq. (\ref{eq:unbias-2}) holds for $T \ge 1$.
Consider $F(h_{1:T+1})$, which can be written as:
\begin{equation} \label{eq:F_T_plus_1}
\begin{split}
&F(h_{1:T+1}) = \underbrace{F(h_{1:T})}_\text{$a_1$}+ \underbrace{\log p_\theta(r_{T+1}|h_{1:T},r_{1:T},h_{T+1})}_\text{$b_1$}+\\
&\underbrace{\sum_{\bar{h}_{T+1}} q_\phi(\bar{h}_{T+1}|h_{1:T},r_{1:T},r_{T+1}) \log \frac{p_\theta(\bar{h}_{T+1}|h_{1:T},r_{1:T})}{q_\phi(\bar{h}_{T+1}|h_{1:T},r_{1:T},r_{T+1})}}_\text{$c_1$}
\end{split}
\end{equation}
where $h_{T+1} \sim q_\phi(h_{T+1}|h_{1:T},r_{1:T},r_{T+1})$.

According to Eq. (\ref{eq:q_phi}), we have
\begin{equation}\label{eq:q_T_plus_1}
q_\phi(h_{1:T+1}|r_{1:T+1})
=q_\phi(h_{1:T}|r_{1:T}) q_\phi(h_{T+1}|h_{1:T},r_{1:T},r_{T+1})
\end{equation}
where we suppress the dependence on $u_t$'s.

Consider $\mathcal{J}_{\text{VL}}(T+1)$, which can be written as:
\begin{equation} \label{eq:J_decompose}
\begin{split}
&\mathcal{J}_{\text{VL}}(T+1)\\
=&\mathbb{E}_{q_\phi(h_{1:T}|r_{1:T})q_\phi(h_{T+1}|h_{1:T},r_{1:T},r_{T+1})} \\
& \left[ \underbrace{\log p_\theta(r_{T+1}|h_{1:T},r_{1:T},h_{T+1})}_\text{$b_2$}+\underbrace{\sum_{t=1}^{T} \log p_\theta(r_{t}|h_{<t},r_{<t},h_t)}_\text{$a_2$} \right] \\
+&\mathbb{E}_{q_\phi(h_{1:T}|r_{1:T})q_\phi(h_{T+1}|h_{1:T},r_{1:T},r_{T+1})} \\
&\left[ \underbrace{\frac{p_\theta({h}_{T+1}|h_{1:T},r_{1:T})}{q_\phi({h}_{T+1}|h_{1:T},r_{1:T},r_{T+1})}}_\text{$c_2$}+\underbrace{\sum_{t=1}^{T} \log \frac{p_\theta(h_{t}|h_{<t},r_{<t})}{q_\phi(h_{t}|h_{<t},r_{<t},r_t)}}_\text{$a_3$}  \right]
\end{split}
\end{equation} 

Next, we will see the equality between $\mathcal{J}_{\text{VL}}(T+1)$ and the expectation over $F(h_{1:T+1})$ under $q_\phi(h_{1:T+1}|u_{1:T+1},r_{1:T+1})$.
\begin{itemize}
    \item The sum of expected $a_2$ and $a_3$ terms in $\mathcal{J}_{\text{VL}}(T+1)$ is $\mathcal{J}_{\text{VL}}(T)$, which equals to the expected $a_1$ term in $F(h_{1:T+1})$, by induction hypothesis;
    \item The expected $b_2$ term in $\mathcal{J}_{\text{VL}}(T+1)$ is exactly the expected $b_1$ term in $F(h_{1:T+1})$;
    \item The expected $c_2$ term in $\mathcal{J}_{\text{VL}}(T+1)$ is exactly the expected $c_1$ term in $F(h_{1:T+1})$.
\end{itemize}
Thereby, we show that the expected $F(h_{1:T+1})$ equals to $\mathcal{J}_{\text{VL}}(T+1)$. This concludes the inductive step.
\end{proof}

\begin{table*}[t]
    \centering
    \includegraphics[width=0.8\linewidth]{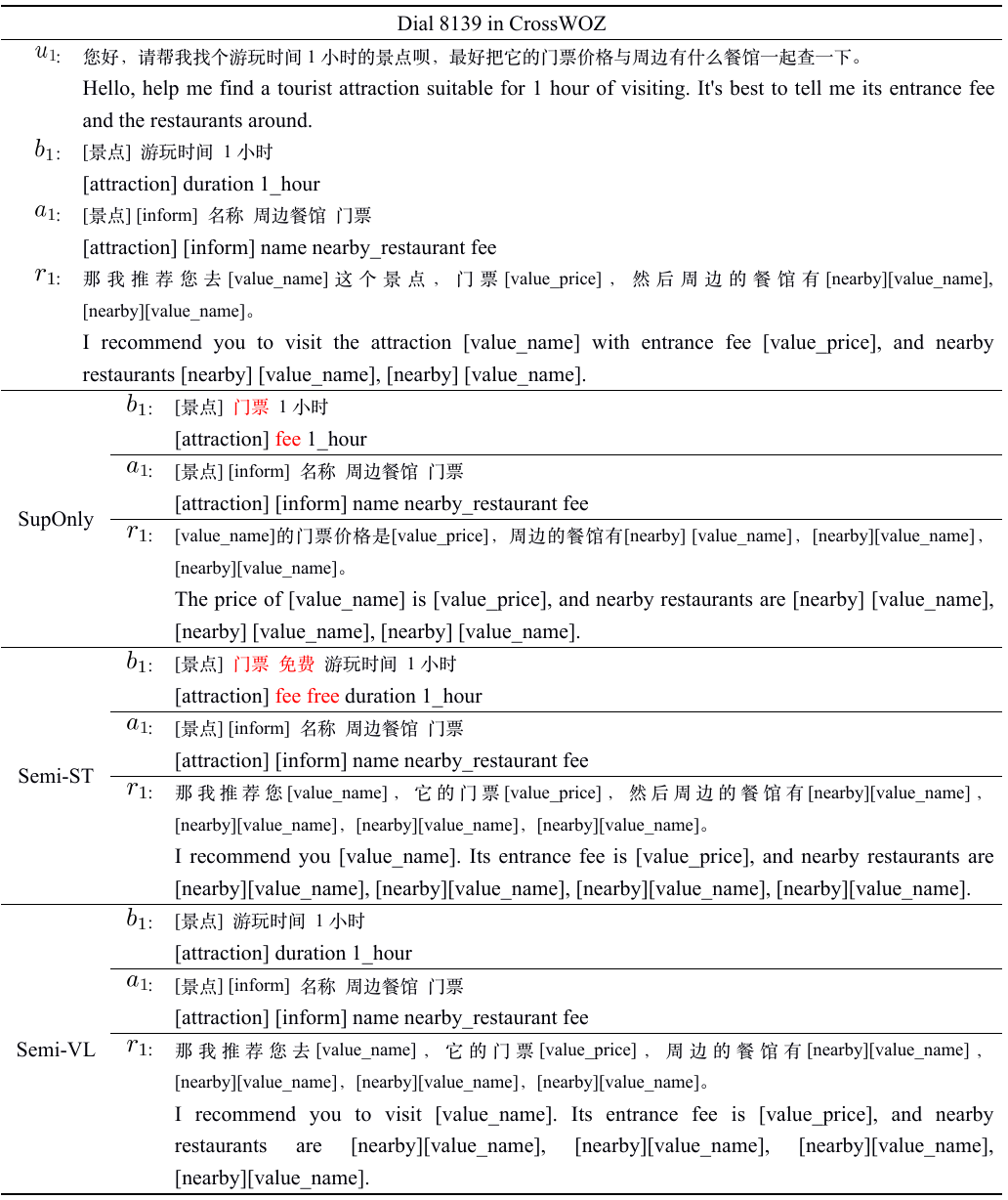}
    \caption{
        An example from CrossWOZ testing set. The label proportion of SupOnly, Semi-ST and Semi-VL is 10\%. For belief states marked in red, SupOnly and Semi-ST generate the wrong slot ``fee'', and SupOnly even generate mismatched slot and value (``fee 1\_hour''). 
    }
    \label{tab:crosswoz}
\end{table*}


\bibliographystyle{IEEEtran}


\bibliography{VLS-GPT}

\begin{thebibliography}{10}
\providecommand{\url}[1]{#1}
\csname url@samestyle\endcsname
\providecommand{\newblock}{\relax}
\providecommand{\bibinfo}[2]{#2}
\providecommand{\BIBentrySTDinterwordspacing}{\spaceskip=0pt\relax}
\providecommand{\BIBentryALTinterwordstretchfactor}{4}
\providecommand{\BIBentryALTinterwordspacing}{\spaceskip=\fontdimen2\font plus
\BIBentryALTinterwordstretchfactor\fontdimen3\font minus
  \fontdimen4\font\relax}
\providecommand{\BIBforeignlanguage}[2]{{%
\expandafter\ifx\csname l@#1\endcsname\relax
\typeout{** WARNING: IEEEtran.bst: No hyphenation pattern has been}%
\typeout{** loaded for the language `#1'. Using the pattern for}%
\typeout{** the default language instead.}%
\else
\language=\csname l@#1\endcsname
\fi
#2}}
\providecommand{\BIBdecl}{\relax}
\BIBdecl

\bibitem{mrkvsic2017neural}
N.~Mrk{\v{s}}i{\'c}, D.~{\'O}. S{\'e}aghdha, T.-H. Wen, B.~Thomson, and
  S.~Young, ``Neural belief tracker: Data-driven dialogue state tracking,'' in
  \emph{Proceedings of the 55th Annual Meeting of the Association for
  Computational Linguistics (ACL)}, 2017.

\bibitem{wen2017latent}
T.~Wen, Y.~Miao, P.~Blunsom, and S.~J. Young, ``Latent intention dialogue
  models,'' in \emph{Proceedings of the 34th International Conference on
  Machine Learning ({ICML})}, D.~Precup and Y.~W. Teh, Eds., 2017.

\bibitem{wen2017a}
T.-H. Wen, D.~Vandyke, N.~Mrk{\v{s}}i{\'c}, M.~Gasic, L.~M.~R. Barahona, P.-H.
  Su, S.~Ultes, and S.~Young, ``A network-based end-to-end trainable
  task-oriented dialogue system,'' in \emph{Proceedings of the 15th Conference
  of the European Chapter of the Association for Computational Linguistics},
  2017.

\bibitem{liu2017end}
B.~Liu and I.~Lane, ``An end-to-end trainable neural network model with belief
  tracking for task-oriented dialog,'' \emph{Proc. Interspeech 2017}, pp.
  2506--2510, 2017.

\bibitem{lei2018sequicity}
W.~{Lei}, X.~{Jin}, M.-Y. {Kan}, Z.~{Ren}, X.~{He}, and D.~{Yin}, ``Sequicity:
  Simplifying task-oriented dialogue systems with single sequence-to-sequence
  architectures,'' in \emph{56th Annual Meeting of the Association for
  Computational Linguistics (ACL)}, 2018.

\bibitem{fsdm}
L.~{Shu}, P.~{Molino}, M.~{Namazifar}, H.~{Xu}, B.~{Liu}, H.~{Zheng}, and
  G.~{Tür}, ``Flexibly-structured model for task-oriented dialogues,'' in
  \emph{Proceedings of the 20th Annual SIGdial Meeting on Discourse and
  Dialogue}, 2019.

\bibitem{zhang2020task}
Y.~{Zhang}, Z.~{Ou}, and Z.~{Yu}, ``Task-oriented dialog systems that consider
  multiple appropriate responses under the same context,'' in \emph{The
  Thirty-Fourth AAAI Conference on Artificial Intelligence (AAAI)}, 2020.

\bibitem{gao2020paraphrase}
S.~Gao, Y.~Zhang, Z.~Ou, and Z.~Yu, ``Paraphrase augmented task-oriented dialog
  generation,'' in \emph{Proceedings of the 58th Annual Meeting of the
  Association for Computational Linguistics}, 2020.

\bibitem{sutskever2014sequence}
I.~Sutskever, O.~Vinyals, and Q.~V. Le, ``Sequence to sequence learning with
  neural networks,'' in \emph{Proc. of Advances in neural information
  processing systems}, 2014.

\bibitem{zhu2006semi}
X.~Zhu, ``Semi-supervised learning literature survey,'' \emph{Technical report,
  University of Wisconsin-Madison}, 2006.

\bibitem{kingma2013auto}
D.~P. Kingma and M.~Welling, ``Auto-encoding variational bayes,'' in \emph{2nd
  International Conference on Learning Representations ({ICLR})}, Y.~Bengio and
  Y.~LeCun, Eds., 2014.

\bibitem{sedst}
X.~{Jin}, W.~{Lei}, Z.~{Ren}, H.~{Chen}, S.~{Liang}, Y.~{Zhao}, and D.~{Yin},
  ``Explicit state tracking with semi-supervision for neural dialogue
  generation,'' in \emph{Proceedings of the 27th ACM International Conference
  on Information and Knowledge Management (CIKM)}, 2018.

\bibitem{zhang-etal-2020-probabilistic}
Y.~Zhang, Z.~Ou, M.~Hu, and J.~Feng, ``A probabilistic end-to-end task-oriented
  dialog model with latent belief states towards semi-supervised learning,'' in
  \emph{Proc. of the Conference on Empirical Methods in Natural Language
  Processing (EMNLP)}, 2020.

\bibitem{radford2018improving}
A.~Radford, K.~Narasimhan, T.~Salimans, and I.~Sutskever, ``Improving language
  understanding by generative pre-training,'' 2018,
  \url{http://openai-assets.s3.amazonaws.com/research-covers/language-unsupervised/language_understanding_paper.pdf}.

\bibitem{devlin2019bert}
J.~Devlin, M.-W. Chang, K.~Lee, and K.~Toutanova, ``{BERT}: Pre-training of
  deep bidirectional transformers for language understanding,'' in \emph{Proc.
  of the Conference of the North American Chapter of the Association for
  Computational Linguistics: Human Language Technologies}, 2019.

\bibitem{heck2020trippy}
M.~Heck, C.~van Niekerk, N.~Lubis, C.~Geishauser, H.-C. Lin, M.~Moresi, and
  M.~Gasic, ``Trippy: A triple copy strategy for value independent neural
  dialog state tracking,'' in \emph{Proceedings of the 21th Annual Meeting of
  the Special Interest Group on Discourse and Dialogue}, 2020.

\bibitem{budzianowski-vulic-2019-hello}
P.~Budzianowski and I.~Vuli{\'c}, ``Hello, it{'}s {GPT}-2 - how can {I} help
  you? towards the use of pretrained language models for task-oriented dialogue
  systems,'' in \emph{Proceedings of the 3rd Workshop on Neural Generation and
  Translation}.\hskip 1em plus 0.5em minus 0.4em\relax Association for
  Computational Linguistics, 2019.

\bibitem{vaswani2017attention}
A.~Vaswani, N.~Shazeer, N.~Parmar, J.~Uszkoreit, L.~Jones, A.~N. Gomez,
  L.~Kaiser, and I.~Polosukhin, ``Attention is all you need,'' in \emph{Proc.
  of Advances in neural information processing systems}, 2017.

\bibitem{radford2019gpt2}
A.~Radford, J.~Wu, R.~Child, D.~Luan, D.~Amodei, and I.~Sutskever, ``Language
  models are unsupervised multitask learners,'' \emph{OpenAI Blog}, vol.~1,
  no.~8, p.~9, 2019.

\bibitem{ham-etal-2020-end}
D.~Ham, J.-G. Lee, Y.~Jang, and K.-E. Kim, ``End-to-end neural pipeline for
  goal-oriented dialogue systems using {GPT}-2,'' in \emph{Proceedings of the
  58th Annual Meeting of the Association for Computational Linguistics (ACL)},
  2020.

\bibitem{hosseini2020simple}
E.~Hosseini-Asl, B.~McCann, C.-S. Wu, S.~Yavuz, and R.~Socher, ``A simple
  language model for task-oriented dialogue,'' \emph{arXiv preprint
  arXiv:2005.00796}, 2020.

\bibitem{peng2020etal}
B.~P.~C. Li, J.~Li, S.~Shayandeh, L.~Liden, and J.~Gao, ``{SOLOIST}: Building
  task bots at scale with transfer learning and machine teaching,''
  \emph{Transactions of the Association for Computational Linguistics (TACL),
  2021}, 2020.

\bibitem{kulhanek2021augpt}
J.~Kulhánek, V.~Hudeček, T.~Nekvinda, and O.~Dušek, ``Au{GPT}: Dialogue with
  pre-trained language models and data augmentation,'' \emph{arXiv preprint
  arXiv:2102.05126}, 2021.

\bibitem{yang2021ubar}
Y.~Yang, Y.~Li, and X.~Quan, ``{UBAR}: Towards fully end-to-end task-oriented
  dialog system with gpt-2,'' in \emph{Proceedings of the AAAI Conference on
  Artificial Intelligence (AAAI)}, 2021.

\bibitem{eric2019multiwoz}
M.~Eric, R.~Goel, S.~Paul, A.~Sethi, S.~Agarwal, S.~Gao, A.~Kumar, A.~K. Goyal,
  P.~Ku, and D.~Hakkani-T{\"u}r, ``Multi{WOZ} 2.1: A consolidated multi-domain
  dialogue dataset with state corrections and state tracking baselines,'' in
  \emph{LREC}, 2020.

\bibitem{zhu2020crosswoz}
Q.~Zhu, K.~Huang, Z.~Zhang, X.~Zhu, and M.~Huang, ``Cross{WOZ}: A large-scale
  chinese cross-domain task-oriented dialogue dataset,'' \emph{Transactions of
  the Association for Computational Linguistics}, vol.~8, pp. 281--295, 2020.

\bibitem{kim2020sequential}
B.~Kim, J.~Ahn, and G.~Kim, ``Sequential latent knowledge selection for
  knowledge-grounded dialogue,'' in \emph{International Conference on Learning
  Representations ({ICLR})}, 2020.

\bibitem{lubis2020lava}
N.~Lubis, C.~Geishauser, M.~Heck, H.-C. Lin, M.~Moresi, C.~van Niekerk, and
  M.~Gasic, ``{LAVA}: Latent action spaces via variational auto-encoding for
  dialogue policy optimization,'' in \emph{Proceedings of the 28th
  International Conference on Computational Linguistics}, 2020, pp. 465--479.

\bibitem{zhao2019rethinking}
T.~{Zhao}, K.~{Xie}, and M.~{Eskenazi}, ``Rethinking action spaces for
  reinforcement learning in end-to-end dialog agents with latent variable
  models,'' in \emph{Annual Conference of the North American Chapter of the
  Association for Computational Linguistics (NAACL-HLT)}, 2019.

\bibitem{bao-etal-2020-plato}
S.~Bao, H.~He, F.~Wang, H.~Wu, and H.~Wang, ``{PLATO}: Pre-trained dialogue
  generation model with discrete latent variable,'' in \emph{Proceedings of the
  58th Annual Meeting of the Association for Computational Linguistics (ACL)},
  2020.

\bibitem{Rezende2014StochasticBA}
D.~J. Rezende, S.~Mohamed, and D.~Wierstra, ``Stochastic backpropagation and
  approximate inference in deep generative models,'' in \emph{ICML}, 2014.

\bibitem{ou2018review}
Z.~Ou, ``A review of learning with deep generative models from perspective of
  graphical modeling,'' \emph{arXiv preprint arXiv:1808.01630}, 2018.

\bibitem{williams1992simple}
R.~J. Williams, ``Simple statistical gradient-following algorithms for
  connectionist reinforcement learning,'' \emph{Machine learning}, vol.~8, no.
  3-4, pp. 229--256, 1992.

\bibitem{jang2016categorical}
E.~Jang, S.~Gu, and B.~Poole, ``Categorical reparameterization with
  gumbel-softmax,'' in \emph{International Conference on Learning
  Representations ({ICLR})}, 2017.

\bibitem{bengio2013estimating}
Y.~Bengio, N.~L{\'e}onard, and A.~Courville, ``Estimating or propagating
  gradients through stochastic neurons for conditional computation,''
  \emph{arXiv preprint arXiv:1308.3432}, 2013.

\bibitem{budzianowski2018large}
P.~Budzianowski, T.-H. Wen, B.-H. Tseng, I.~Casanueva, U.~Stefan, R.~Osman, and
  M.~Ga{\v{s}}i\'c, ``Multi{WOZ} - a large-scale multi-domain wizard-of-oz
  dataset for task-oriented dialogue modelling,'' in \emph{Proceedings of the
  2018 Conference on Empirical Methods in Natural Language Processing (EMNLP)},
  2018.

\bibitem{mehri2019structured}
S.~{Mehri}, T.~{Srinivasan}, and M.~{Eskenazi}, ``Structured fusion networks
  for dialog,'' in \emph{Proceedings of the 20th Annual SIGdial Meeting on
  Discourse and Dialogue}, 2019.

\bibitem{gillick1989some}
L.~Gillick and S.~J. Cox, ``Some statistical issues in the comparison of speech
  recognition algorithms,'' in \emph{International Conference on Acoustics,
  Speech, and Signal Processing,}.\hskip 1em plus 0.5em minus 0.4em\relax IEEE,
  1989, pp. 532--535.

\bibitem{machavcek2020elitr}
D.~Mach{\'a}{\v{c}}ek, J.~Kratochv{\'\i}l, S.~Sagar, M.~{\v{Z}}ilinec,
  O.~Bojar, T.-S. Nguyen, F.~Schneider, P.~Williams, and Y.~Yao, ``{ELITR}
  non-native speech translation at {IWSLT} 2020,'' \emph{arXiv preprint
  arXiv:2006.03331}, 2020.

\end{thebibliography}

\begin{IEEEbiography}[{\includegraphics[width=1in,height=1.25in,clip,keepaspectratio]{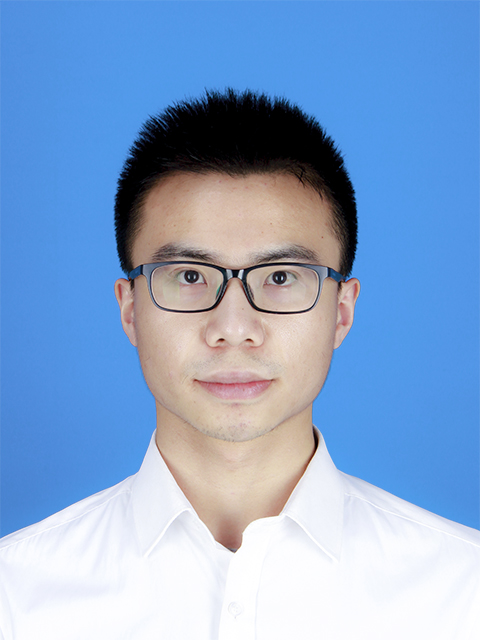}}]{Hong Liu} received the B.S. degree in electronic engineering from Tsinghua University in 2021. He is currently pursuing a master degree at the Department of Electronic Engineering of Tsinghua University under the supervision of Zhijian Ou. His recent research interests focus on semi-supervised learning and reinforcement learning in dialogue systems.
\end{IEEEbiography}
\vspace{-2em}

\begin{IEEEbiography}[{\includegraphics[width=1in,height=1.25in,clip,keepaspectratio]{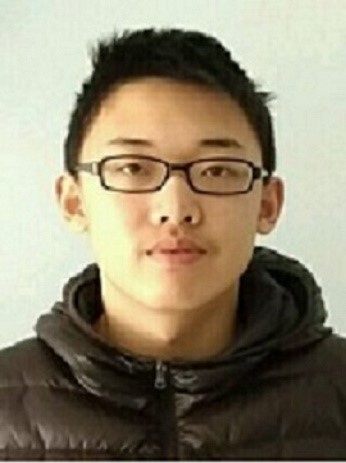}}]{Yucheng Cai} received the B.S. degree in electronic engineering from Tsinghua University in 2022. Since 2022, he has been pursuing a master degree at the Department of Electronic Engineering in Tsinghua University under the supervision of Zhijian Ou. His recent research interests focus on building a high quality chatbot for real-life scenarios and semi-supervised learning theory.
\end{IEEEbiography}
\vspace{-2em}

\begin{IEEEbiography}[{\includegraphics[width=1in,height=1.25in,clip,keepaspectratio]{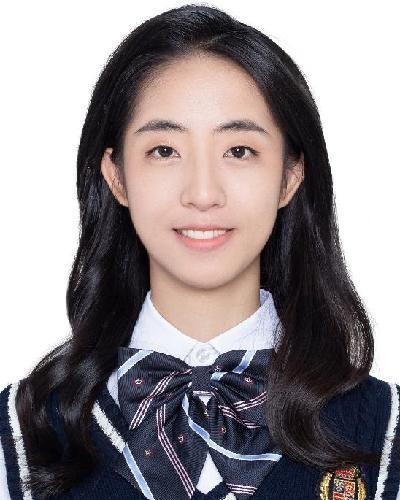}}]{Zhenru Lin} received the B.S. degree in electronic engineering from Tsinghua University in 2022. In the summer of 2021, she worked as an intern under the supervision of Zhijian Ou. Since 2022, she has been pursuing a Ph.D. degree at the Institute for Interdisciplinary Information of Tsinghua University.
\end{IEEEbiography}
\vspace{-2em}

\begin{IEEEbiography}[{\includegraphics[width=1in,height=1.25in,clip,keepaspectratio]{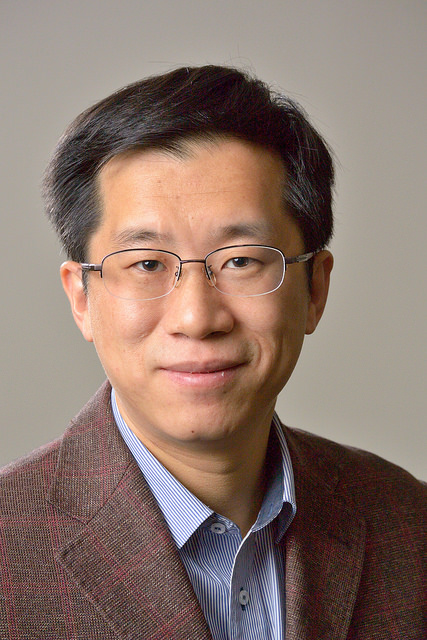}}]{Zhijian Ou} (Senior Member, IEEE) is an associate professor with the Department of Electronic Engineering at Tsinghua University. He received his Ph.D. from Tsinghua University in 2003. He currently serves as Associate Editor of “IEEE/ACM Transactions on Audio, Speech and Language Processing”, Editorial Board Member of “Computer Speech and Language”, member of IEEE Speech and Language Processing Technical Committee, and was General Chair of SLT 2021, EMNLP 2022 SereTOD workshop, and Tutorial Chair of INTERSPEECH 2020. He conducts research in the general area of speech and language processing (particularly speech recognition, dialogue systems) and machine intelligence (particularly with graphical models and deep learning).
\end{IEEEbiography}
\vspace{-2em}

\begin{IEEEbiography}[{\includegraphics[width=1in,height=1.25in,clip,keepaspectratio]{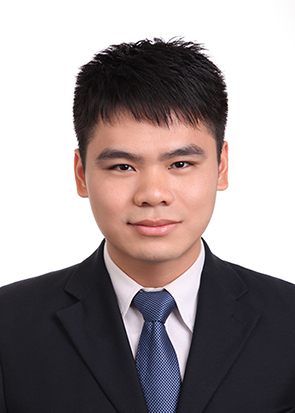}}]{Yi Huang} is currently a Senior Researcher at the AI and Intelligent Operation Center, China Mobile Research Institute. His research interests are dialogue systems and knowledge engineering. He is a frequent reviewer and organizer for major natural language international conferences and journals such as ACL, EMNLP, AAAI and IJCAI. He has over 20 professional publications and 30 patents.
\end{IEEEbiography}
\vspace{-2em}

\begin{IEEEbiography}[{\includegraphics[width=1in,height=1.25in,clip,keepaspectratio]{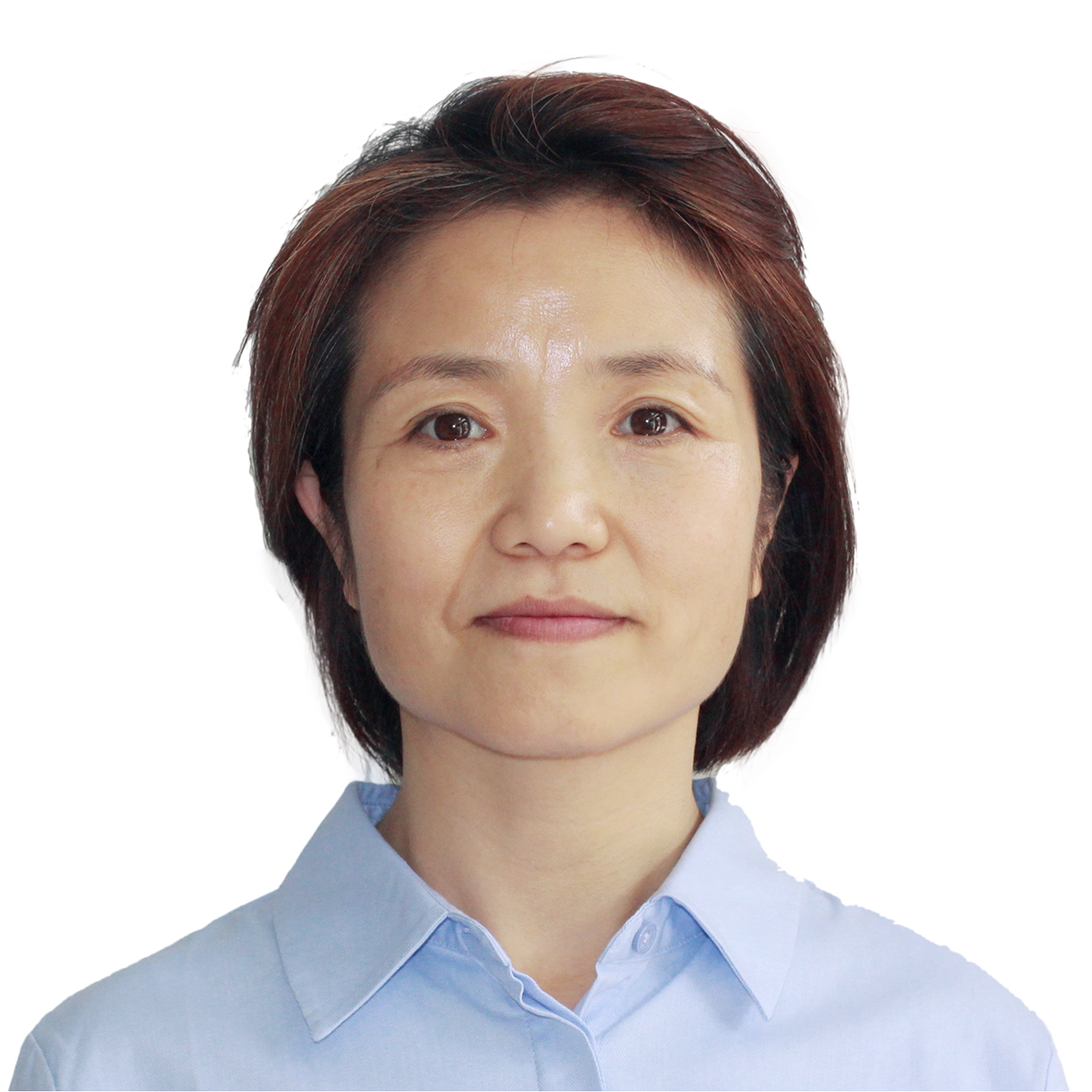}}]{Junlan Feng} (Fellow, IEEE) Chief Scientist at China Mobile Research, Board Chair of Linux Foundation Network. Dr. Feng received her Ph.D. from Chinese Academy of Sciences, and joined AT\&T Labs Research in 2001 as a principal researcher on speech recognition, language understanding and data mining until 2013. She has led the R\&D team on artificial intelligence at China Mobile since then.  Dr. Feng has over 100 publications and over 60 issued patents.
\end{IEEEbiography}

\end{document}